\documentclass[10pt,twocolumn,letterpaper]{article}
\usepackage{cvpr}
\usepackage{multirow}
\usepackage{pifont}
\usepackage{xspace}
\usepackage{graphicx}
\usepackage{amsmath}
\usepackage{wrapfig}
\usepackage{xcolor}
\usepackage{colortbl}

\usepackage{caption}
\usepackage{etoc}
\etocdepthtag.toc{mtchapter}
\etocsettagdepth{mtchapter}{subsection}
\etocsettagdepth{mtappendix}{none}

\newcommand{\framework}{\textsc{CineNeuron}\xspace}
\newcommand{\frameworkplain}{\textsc{CineNeuron}\xspace}
\newcommand{\frameworkNoSpace}{\textsc{CineNeuron}}

\newlength\savewidth

\newcommand{\tabincell}[2]{\begin{tabular}{@{}#1@{}}#2\end{tabular}}

\definecolor{darkred}{rgb}{0.7,0.1,0.1}
\definecolor{darkgreen}{rgb}{0.1,0.6,0.1}
\newcommand{\cmark}{\color{darkgreen}{\ding{51}}}

\usepackage{amsmath,amsfonts,bm}

\def\eqref#1{equation~\ref{#1}}

\def\1{\bm{1}}

\DeclareMathAlphabet{\mathsfit}{\encodingdefault}{\sfdefault}{m}{sl}
\SetMathAlphabet{\mathsfit}{bold}{\encodingdefault}{\sfdefault}{bx}{n}

\definecolor{cvprblue}{rgb}{0.21,0.49,0.74}
\usepackage[pagebackref,breaklinks,colorlinks,allcolors=cvprblue]{hyperref}
\usepackage[accsupp]{axessibility}

\def\confName{CVPR}
\def\confYear{2026}

\title{Bridging Brain and Semantics: A Hierarchical Framework for Semantically Enhanced fMRI-to-Video Reconstruction}

\author{
Yujie Wei$^{1,*}$, Chenglong Ma$^{1,*}$, Jianxiong Gao$^1$, Chenhui Wang$^1$, Shiwei Zhang$^2$, \\ Biao Gong$^3$, Shuai Tan$^3$, Hangjie Yuan$^2$, Hongming Shan$^{1, \dagger}$
 \vspace{0.6em} \\
 $^1$Fudan University \qquad $^2$Alibaba Group \qquad $^3$Ant Group 
 \\ 
 {\tt\small yjwei22@m.fudan.edu.cn},\quad
 {\tt\small clma24@m.fudan.edu.cn}
}

\usepackage[misc]{ifsym}
\newcommand\blfootnote[1]{
    \begingroup
    \renewcommand\thefootnote{}\footnote{#1}
    \addtocounter{footnote}{-1}
    \endgroup
}
\usepackage[hang]{footmisc}

\begin{document}
\maketitle

{
\blfootnote{
$^*$Equal Contribution\quad$^\dagger$Corresponding Author \\
}
}

\begin{abstract}
Reconstructing dynamic visual experiences as videos from 
functional magnetic resonance imaging (fMRI)
is pivotal for advancing the understanding of neural processes. 
However,
current fMRI-to-video reconstruction methods are hindered by a semantic gap between noisy fMRI signals and the rich content of videos, 
stemming from a reliance on incomplete semantic embeddings that neither capture video-specific cues (\eg, actions) nor integrate prior knowledge.
To this end, we draw inspiration from the dual-pathway processing mechanism in human brain 
and introduce \framework, a novel hierarchical framework for semantically enhanced video reconstruction from fMRI signals with two synergistic stages. 
First, a bottom-up semantic enrichment stage maps fMRI signals to a rich embedding space that comprehensively captures textual semantics, image contents, action concepts, and object categories. 
Second, a top-down memory integration stage utilizes the proposed Mixture-of-Memories method to dynamically select relevant ``memories'' from previously seen data and fuse them with the fMRI embedding to refine the video reconstruction.
Extensive experimental results on two fMRI-to-video benchmarks demonstrate that \frameworkplain surpasses state-of-the-art methods across various metrics. 
\end{abstract}

\section{Introduction}
Understanding how the human brain processes visual information has long been a central goal in cognitive neuroscience~\cite{esteban2019fmriprep, tang2023semantic, allen2022massive, daly2023neural, li2024visual}. Among various efforts, reconstructing visual experiences from neural signals, particularly functional magnetic resonance imaging (fMRI), offers valuable insights into the neural representations underlying visual perception. 
While earlier studies~\cite{chen2023seeing, scotti2023reconstructing, li2024neurobolt, bai2023dreamdiffusion} have made substantial progress in reconstructing static images from fMRI signals, extending this success to continuous visual stimuli (\eg, videos) remains an open challenge. 

\begin{figure}[t]
  \centering
  \includegraphics[width=1.0\linewidth]{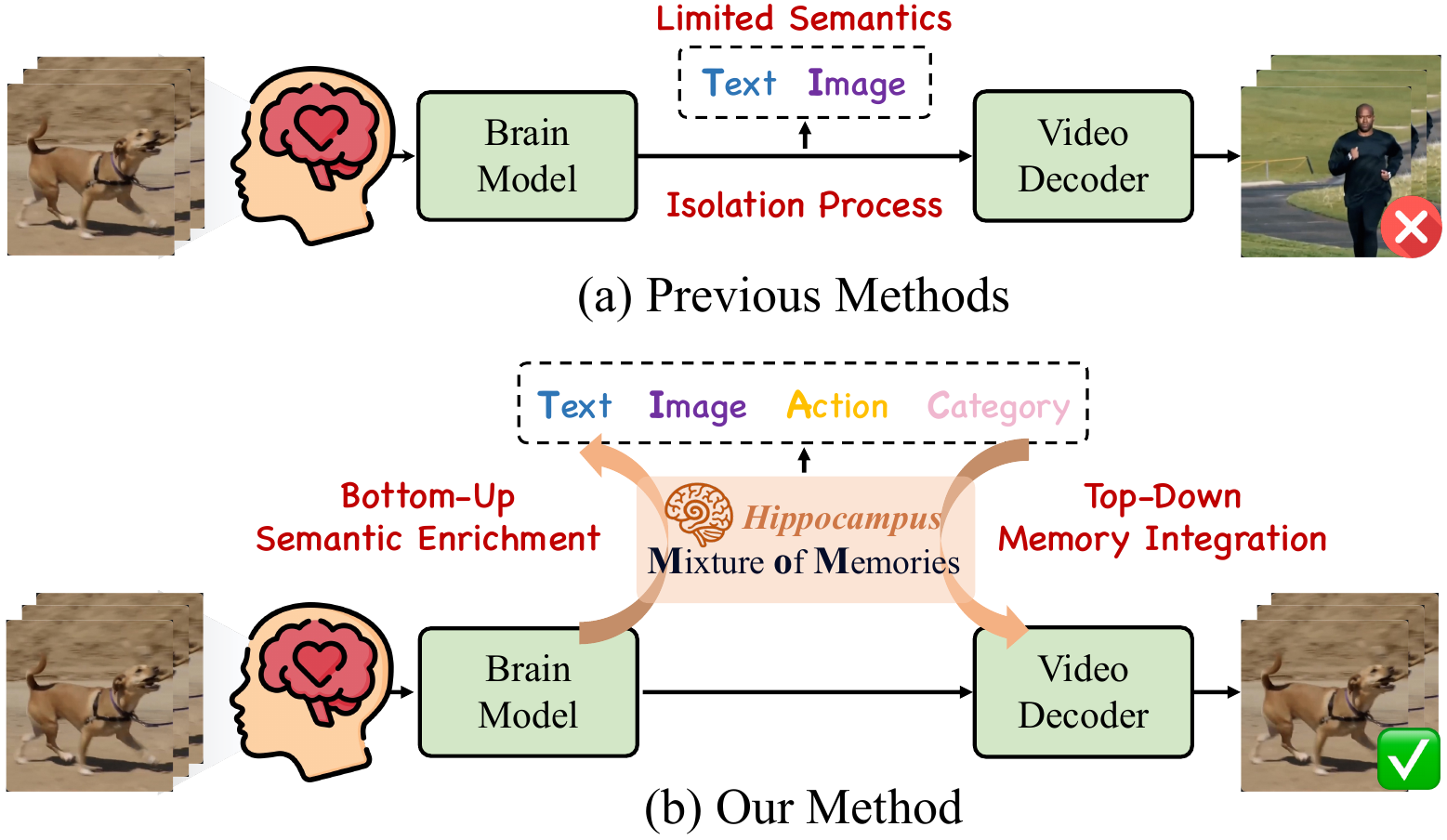}
  \caption{\textbf{Comparison with previous methods.} (a) Previous methods often align fMRI embeddings with limited semantics in an isolation process, relying only on the current stimulus and yielding semantically inaccurate results. (b) Our method enriches the fMRI embeddings with comprehensive video semantics and introduces Mixture-of-Memories to dynamically select and fuse prior knowledge, producing semantically coherent videos.
  }
  \label{fig:motivation}
  \vspace{-5mm}
\end{figure}

The difficulty mainly stems from accurately recovering both spatial content and temporal dynamics of videos from raw fMRI signals that have a low signal-to-noise ratio and limited temporal resolution~\cite{baijot2021signal}.
The inherent sparsity and noise of fMRI further blur the high-level meaning of the underlying neural responses~\cite{ozcelik2023natural, yang2024brainmae},
making it difficult to extract comprehensive semantics from fMRI data alone. 
Prior works~\cite{chen2023cinematic, gong2024neuroclips, lu2024animate} have attempted to enrich semantics by aligning fMRI embeddings with additional images and text representations.
However, as shown in Fig.~\ref{fig:motivation}, these methods still suffer from low semantic coherence and reconstruct videos with incorrect objects for two reasons: 
\textbf{1)} They ignore the rich semantics embedded in video stimuli, such as actions and object categories, which are essential for faithful video reconstruction;
\textbf{2)} 
They treat the fMRI reconstruction in isolation, using only the current stimulus.
In fact, the human brain interprets the visual world by incorporating both incoming stimuli and memories of learned semantic concepts~\cite{takehara2014entorhinal, spens2024generative}.
Therefore, \textit{enriching the semantic capacity of fMRI embeddings and integrating prior knowledge are both crucial for bridging the semantic gap between brain activity and reconstructed videos.
}

This hierarchical design principle is also consistent with the brain's dual processing pathways~\cite{gilbert2013top, de2018expectations}. Specifically, a bottom-up pathway extracts high-level semantics by accumulating sensory evidence from early to higher-order visual regions, while a top-down pathway delivers semantic predictions and integrated memories from hippocampal systems back to the sensory cortex to refine stimulus perception. This synergistic interaction is fundamental to how the brain constructs a coherent perception of the real world.

Drawing inspiration from this dual-pathway mechanism, we present \frameworkplain, a hierarchical framework for semantically enhanced fMRI-to-video reconstruction by unifying two synergistic stages:
\textbf{1)} \emph{bottom-up semantic enrichment}, 
which enriches the fMRI representations with multimodal semantics extracted from the current video stimuli, and 
\textbf{2)} \emph{top-down memory integration}, 
which incorporates ``memories'' from the semantic embeddings of previously seen data to supplement and refine the fMRI representations, ensuring a coherent video reconstruction.

Specifically, in the first stage, we decompose the semantics hidden in the video stimuli into multiple dimensions 
and extract corresponding embeddings using heterogeneous pre-trained encoders. 
Beyond aligning fMRI only with image and text modalities, we design explicit video tasks using action concepts and object categories.
Then, we train a Brain Model to map fMRI signals into a space capturing these comprehensive semantics, creating a semantically rich foundation for fMRI-to-video reconstruction. 

In the second stage, we introduce a Mixture-of-Memories (MoM) module to refine the Stage-1 fMRI embedding by fusing it with prior ``memories,'' \ie, multimodal embeddings from previously seen data.
Analogous to hippocampal processing, MoM proceeds in two coordinated steps: retrieval and integration.
The retrieval step uses a modality-aware router to dynamically assign weights to multimodal memories and selects the most relevant ones via weighted similarity.
The integration step then aggregates these selected memories with the fMRI embedding, producing a fused representation that incorporates prior knowledge while preserving current neural evidence. 
This final representation is then passed to a video decoder to reconstruct temporally consistent and semantically coherent videos.

We evaluate the effectiveness of \frameworkplain on two challenging fMRI-to-video reconstruction datasets, cc2017~\cite{wen2018neural} and CineBrain~\cite{gao2025cinebrain}, where it outperforms state-of-the-art methods across metrics and reconstructs semantically accurate and visually coherent videos.

Our contributions are summarized as follows:
\textbf{1)} We present \frameworkplain, a novel hierarchical framework for semantically enhanced fMRI-to-video reconstruction by synergizing the bottom-up semantic enrichment with the top-down memory integration.
\textbf{2)} 
We propose to enrich the semantics of fMRI signals by learning explicit tasks for video stimuli, 
capturing comprehensive semantics with image, text, action, and category concepts.
\textbf{3)}
We propose a novel Mixture-of-Memories method that dynamically selects and fuses relevant multimodal embeddings with the fMRI embedding to refine video decoding.
\textbf{4)} 
Extensive experimental results on two fMRI-to-video benchmarks demonstrate the superior performance of \frameworkplain over state-of-the-art methods.
\section{Related Work}

\subsection{Brain Decoding} 
Decoding visual stimuli from brain activity, particularly fMRI signals, has attracted growing attention in recent years~\cite{chen2023seeing,scotti2023reconstructing,jiao2019decoding,quan2024psychometry,wang2024mindbridge,benchetrit2023brain, wang2025zebra, lu2025cognitive, liu2023brainclip, jing2025pinpointing, jing2025beyond, liu2025see, ma2024hierarchical, jing2026evoke, jing2026damind, jing2026mind, gao2023mind3d, gao2025mind3d++, diakite2025dual}. While previous works have shown promising results in decoding static images from fMRI signals~\cite{zheng2020decoding, xie2024brainram, zhao2025memory, gong2025mindtuner, yeung2025reanimating}, the human visual perception is inherently dynamic and continuous. This gap highlights the need for fMRI-to-video reconstruction, which enables a deeper understanding of how the brain processes real-world stimuli. 
However, reconstructing video content from fMRI signals presents greater challenges due to the low temporal resolution and signal-to-noise ratio of the measurements~\cite{wang2022reconstructing, lu2024animate, wang2025neurons, liu2024eeg2video, fosco2024brain}. 

Recent fMRI-to-video reconstruction methods typically learn semantic embedding from raw fMRI signals and then drive a video generator~\cite{lu2024animate, wang2025neurons, zeng2025mindshot}.
For instance,
Mind-Video~\cite{chen2023cinematic} aligns a self-supervised fMRI encoder to CLIP~\cite{radford2021clip} and adapts an inflated Stable Diffusion~\cite{rombach2022high} model for video reconstruction. NeuroClips~\cite{gong2024neuroclips} advances this further by encoding semantic keyframes from low-level perceptual flows, improving the smoothness of generated videos. Despite these advances, previous methods capture only shallow semantics as they rely primarily on image-text aligned spaces and treat reconstruction as an isolated process, using only the current fMRI stimulus and ignoring prior learned knowledge. 
In contrast, our work explicitly enriches fMRI with comprehensive semantics by learning tasks over text, image, \emph{action}, and \emph{category} concepts and integrating multimodal memories for refined decoding.

\subsection{Diffusion Models for Video Generation}
Diffusion models have become the dominant paradigm for image and video synthesis~\cite{ho2020denoising, animatediff, wei2025routing, han2025turning, han2025can, MIR-2025-02-045}, primarily due to their superior generation fidelity and flexible conditioning mechanisms~\cite{wei2024dreamvideo, wei2024dreamvideo2, tan2025edtalk, AnimateX2025, tan2023emmn, tan2025fixtalk, Li_2025_ICCV, li2026flashmotion}.
Early works like VDM~\cite{VDM} pioneer this direction by extending image diffusion to videos, while subsequent text-to-video models~\cite{modelScope, chen2023videocrafter1, chen2024videocrafter2} integrate spatiotemporal modules for higher temporal consistency.
Driven by the rapid progress of Transformers~\cite{vaswani2017attention} in understanding~\cite{liu2025hybrid, liu2025capability} and generation~\cite{genmo2024mochi, wei2025dreamrelation, liu2025showtable} tasks, Diffusion Transformers (DiT)~\cite{DiT} establishes a new paradigm that further enhances generation quality~\cite{kong2024hunyuanvideo, wei2026dreamvideo-omni}.
Modern video generation models push long-range coherence with advanced DiT architectures.
For example, 
CogVideoX~\cite{yang2024cogvideox} introduces a spatiotemporal VAE~\cite{kingma2013auto, chen2025masked} and a multimodal DiT with full attention.
Wan~\cite{wan2025wan} adopts a cross-attention DiT trained on large-scale data, enhancing video coherence.
In this work, we adopt Wan2.1 1.3B as our default video decoder backbone, but our framework is model-agnostic and can be readily integrated into other models like CogVideoX.

\section{Method}
\begin{figure*}[t]
  \centering
  \includegraphics[width=1.0\linewidth]{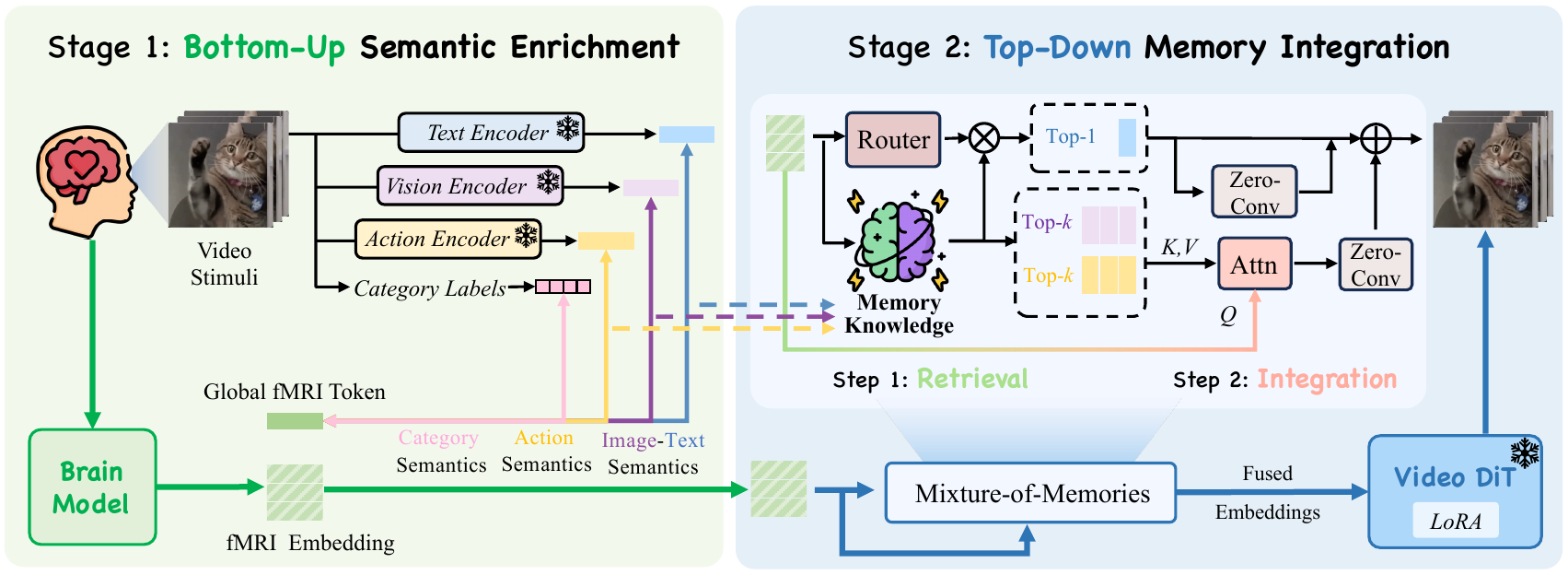}
  \caption{\textbf{Overview of the proposed \framework.} In stage 1, given an input fMRI-video pair, the fMRI signals are first embedded by a Brain Model and enriched with the text, image, action, and category semantics extracted from the video. In stage 2, the proposed Mixture-of-Memories method dynamically selects multimodal embeddings from previously seen data via a router and fuses them with the fMRI embeddings via a fusion mechanism, guiding the Video DiT for semantically-enhanced and high-quality video reconstruction.
  }
  \label{fig:framework}
\end{figure*}

The overall framework of \frameworkplain is illustrated in Fig.~\ref{fig:framework}, which consists of two main learning stages. 
\textit{1)} In the bottom-up semantic enrichment stage (Sec.~\ref{sec:stage1}), 
a Brain Model is designed to map fMRI signals to an embedding space that captures comprehensive semantics. 
\textit{2)} In the top-down memory integration stage (Sec.~\ref{sec:stage2}), 
the proposed Mixture-of-Memories module first utilizes fMRI embeddings to retrieve relevant text, image, and action embeddings from an established memory pool, and then
integrate them into the fMRI embeddings to serve as the input condition for the video decoder.

\subsection{Bottom-Up Semantic Enrichment}
\label{sec:stage1}
Extracting reliable semantic cues from noisy raw fMRI signals remains a bottleneck in reconstructing dynamic videos, which necessitates transforming the fMRI signals into more compact, informative representations using multimodal semantic priors~\cite{MIR-2022-12-391,gao2025multi}.
To address this, we enrich the fMRI embeddings with comprehensive semantics, including text, image, action, and object categories, all of which are crucial for video reconstruction. This is driven by a suite of dedicated learning tasks: \emph{image-text semantic alignment}, \emph{action semantic alignment}, and \emph{category semantic learning}.

To begin with, we build on the previous works~\cite{chen2023cinematic, li2024enhancing, gao2025cinebrain} to design a transformer-based Brain Model that produces two outputs: a global fMRI token summarizing the fMRI signal and an fMRI embedding that retains finer contextual patterns. 
This dual output allows for functional decoupling, where the global fMRI token supports alignment with external semantic representations and the fMRI embedding facilitates downstream feature fusion and video reconstruction in the subsequent stage.
Below, we detail the semantic enrichment within each task.

\noindent\textbf{Image-Text Semantic Alignment.}\quad
Given a video clip $ \bm{y} \in \mathbb{R}^{B \times F \times C \times H \times W} $ and the paired fMRI signals $\bm{x} \in \mathbb{R}^{B \times N}$, 
we first feed $\bm{x}$ into the Brain Model $f$ to obtain the fMRI global token $\bm{f^c} \in \mathbb{R}^{B \times D}$ and the fMRI embedding $\bm{f^e} \in \mathbb{R}^{B \times L \times D^{\prime}}$. 
The video $\bm{y}$ and its caption $c$
are processed by the CLIP encoder to produce multi-frame image embeddings $\bm{e^\text{img}} \in \mathbb{R}^{BF \times D}$ and text embeddings $\bm{e^\text{txt}} \in \mathbb{R}^{B \times D}$, respectively. 
Then, a projection head $\varphi_v$ aggregates the multi-frame image embeddings into a consolidated image embedding $\bm{\hat{e}^\text{img}} = \varphi_v(\bm{e^\text{img}}) \in \mathbb{R}^{B \times D}$.
Finally, we employ the InfoNCE loss~\cite{oord2018representation} to align the global fMRI token with image and text embeddings:
\begin{align}
\mathcal{L}_{\text {clip}} &= \mathcal{L}_{\text {info}} (\bm{f^c}, \bm{\hat{e}^\text{img}}) + \mathcal{L}_{\text {info}} (\bm{f^c}, \bm{e^\text{txt}}),
\\
\mathcal{L}_{\text {info}} (\bm{f^c}, \bm{e}) &= -\frac{1}{B} \sum_{i=1}^B \log \frac{\exp \left(\bm{f^c}_i \cdot \bm{e}_i / \tau\right)}{\sum_{j=1}^B \exp \left(\bm{f^c}_i \cdot \bm{e}_j / \tau\right)}, 
\label{eq:infoNCE}
\end{align}
where $\tau$ is the temperature.

\noindent\textbf{Action Semantic Alignment.}\quad
Apart from image and text representations, actions such as walking and swimming are vital to understanding video content and preventing semantic discrepancies for the fMRI-to-video reconstruction. 
To this end, we introduce an action alignment task using ViCLIP~\cite{wang2023internvid}, a video understanding model pretrained on a large-scale dataset of 10 million video-text pairs. ViCLIP exhibits robust zero-shot capabilities for video understanding and action recognition, thereby effectively capturing the action and temporal information in the input videos~\cite{huang2024vbench}. 
We process the video $\bm{y}$ with ViCLIP to obtain action embeddings $\bm{e^\text{act}}$, and map the fMRI global token into the action space using an action head $\varphi_a$, yielding $\bm{f^a} = \varphi_a(\bm{f^c})$. 
Consequently, we compute the action contrastive loss $\mathcal{L}_{\text {action}}$ to align the projected fMRI token and the action embedding, formulated as  $\mathcal{L}_{\text {action}} = \mathcal{L}_{\text {info}} (\bm{f^a}, \bm{e^\text{act}})$.

\noindent\textbf{Category Semantic Learning.}\quad
While contrastive learning ensures semantic alignment at the embedding level, explicitly identifying objects within a video helps the Brain Model capture category semantics. 
To achieve this, we introduce a multi-label classification task. 
Specifically, we utilize Qwen2.5-VL~\cite{wang2024qwen2} to extract object categories from video captions based on the MSCOCO category list~\cite{lin2014microsoft}, and train a classification head $\varphi_c$ for category prediction using binary cross-entropy (BCE) loss. 
However, directly training the Brain Model with this objective faces two main challenges: 
\textit{1)} the large number of categories complicates the task, impeding effective category semantic learning; and 
\textit{2)} the class imbalance can bias predictions towards frequent classes. 
To address these issues, we simplify the categories by filtering and merging infrequent classes into a reduced set of superclasses, and apply Focal Loss~\cite{lin2017focal} to dynamically reweight samples within each category. The final classification loss $\mathcal{L}_{\text {cls}}$ is a combination of BCE and focal loss. 
Finally, the loss for this stage is given by:
\begin{equation}
\mathcal{L}_{\text {stage1}} = \mathcal{L}_{\text {clip}} + \lambda_1 \mathcal{L}_{\text {action}} + \lambda_2 \mathcal{L}_{\text {cls}}.
\end{equation}

\subsection{Top-Down Memory Integration}
\label{sec:stage2}
The first stage equips fMRI embeddings with multimodal semantic information, yet further refinement is necessary to enhance and supplement semantic and perceptual details in the sparse fMRI signals.
Inspired by the hippocampal memory replay, we introduce Mixture-of-Memories (MoM), a two-step retrieval-and-integration method. MoM first replays and retrieves multimodal embeddings from a memory pool built from seen training videos, and then integrates these memories into the fMRI embedding to guide the decoder, refining reconstructed videos.

\noindent\textbf{Retrieval Step.}\quad 
The first step of MoM is to retrieve the most relevant memories using the fMRI embeddings from the bottom-up semantic enrichment stage.
A straightforward solution is to compare the text embeddings in the memory pool with the fMRI embeddings, 
obtain the most similar one as a condition, and feed it into the subsequent Video DiT model for reconstruction.
However, such na\"ive strategy is sensitive to imprecise retrieval due to noise or ambiguity in the fMRI signals.

To this end, we incorporate multimodal embeddings as memories and 
adaptively weighs the contributions of each modality.
Let $\mathcal{M}$ denote the memory pool with $N$ entries, where each entry $i$ is a tuple of tri-modality embeddings $(\bm{e}^{\text{txt}}_{i}, \bm{e}^{\text{img}}_{i}, \bm{e}^{\text{act}}_{i})$ extracted from the same training video.
A routing network $R$ first computes instance-specific retrieval weights of three modalities using fMRI embedding $\bm{f^e}$:
\begin{equation}
    W_r = [w^\text{txt}, w^\text{img}, w^\text{act}] = \mathrm{softmax}(R(\bm{f^e})).
\end{equation}
Instead of retrieving from each modality independently, we compute a \textit{mixture score} $S_i$ for each memory entry through a weighted sum of its multimodal similarities to $\bm{f^e}$:
\begin{equation}
\label{eq:unified_score}
S_i = \sum_{m \in \mathcal{M}_{\text{modal}}} w_{m} \cdot \mathrm{sim}(\bm{f^e}, \bm{e}^{m}_{i}), \quad i \in \{1, \ldots, N\},
\end{equation}
where $\mathcal{M}_{\text{modal}} = \{\text{txt}, \text{img}, \text{act}\}$ and $\mathrm{sim}(\cdot, \cdot)$ denotes cosine similarity. 
We then re-rank all memory entries based on their mixture scores $S_i$. Specifically, let $\{i_1, i_2, \ldots, i_K\}$ be the indices of the memory entries corresponding to the top-$K$ highest scores, we select the top-$1$ text embedding $\bm{e}^{\text{mem}}_{\text{txt}} = \bm{e}^{\text{txt}}_{i_1}$ from the highest-scoring entry, along with the sets of top-$K$ image and action embeddings $\bm{e}^{\text{mem}}_{\text{img}} = \{\bm{e}^{\text{img}}_{i_k}\}_{k=1}^K$ and $\bm{e}^{\text{mem}}_{\text{act}} = \{\bm{e}^{\text{act}}_{i_k}\}_{k=1}^K$.
This mixture weighting strategy allows the model to dynamically prioritize the most relevant modality for each input instance, enhancing retrieval precision.

\noindent\textbf{Integration Step.}\quad
The second step of MoM is to fuse retrieved multimodal embeddings from the memory pool into the fMRI embedding.
Specifically, we design a fusion mechanism using two cross-attention layers and zero-convolution layers. 
In cross-attention layers, the fMRI embedding $\bm{f^e}$ functions as the query, while $K$ retrieved image embeddings $\bm{e}^{\text{mem}}_{\text{img}}$ and action embeddings $\bm{e}^{\text{mem}}_{\text{act}}$ serve as keys and values, respectively.
This attention mechanism injects visual and action cues into 
the fMRI embedding, producing an enhanced representation $\bm{\hat{f}^e}$:
\begin{equation}
\begin{aligned}
    \bm{\hat{f}^e} &= \mathrm{CrossAttention}(\bm{Q^e}, \bm{K^\text{img}}, \bm{V^\text{img}}) \\ 
    &+ \mathrm{CrossAttention}(\bm{Q^e}, \bm{K^\text{act}}, \bm{V^\text{act}}),
\end{aligned}
\end{equation}
where $\bm{Q^e} = \bm{W_Q}\bm{f^e}$, $\bm{K^\text{img}} = \bm{W_K}\bm{e}^{\text{mem}}_{\text{img}}$, and $\bm{V^\text{img}} = \bm{W_V}\bm{e}^{\text{mem}}_{\text{img}}$.
$\bm{K^\text{act}}$ and $\bm{V^\text{act}}$ are computed similarly. 

We then employ a dual-stream structure to fuse $\bm{\hat{f}^e}$ with the retrieved top-$1$ text embedding $\bm{e}^{\text{mem}}_{\text{txt}}$.
Concretely, $\bm{\hat{f}^e}$ is first passed through a normalization layer~\cite{xie2024sana, tan2024mimir} to scale it comparably to the text embeddings, facilitating effective integration. 
Following normalization, a zero-conv layer $\mathcal{Z}_\text{fMRI}$ is applied to $\bm{\hat{f}^e}$ to ensure stability in training by initializing outputs to zero.
To balance the contributions of both modalities, another zero-conv layer $\mathcal{Z}_\text{txt}$ is added for text embedding $\bm{e}^{\text{mem}}_{\text{txt}}$ in a residual manner, ensuring the initial fused embedding equates to $\bm{e}^{\text{mem}}_{\text{txt}}$:
\begin{equation}
\begin{aligned}
\bm{z_f} =\mathcal{Z}_\text{fMRI}(\text{Norm}(\bm{\hat{f}^e})),\quad
\bm{z_t} =\mathcal{Z}_\text{txt}(\bm{e}^{\text{mem}}_{\text{txt}}) + \bm{e}^{\text{mem}}_{\text{txt}}.
\end{aligned}
\end{equation}
Finally, the embeddings are then summed to produce the fused embeddings $\bm{f^{\text{fuse}}}$ as follows:
\begin{equation}
\begin{aligned}
\bm{f^{\text{fuse}}} = \bm{z_t} + \alpha * \bm{z_f},
\label{eq:fuse_embed}
\end{aligned}
\end{equation}
where $\alpha$ is the weighting factor for balance.
The final fused embedding $\bm{f^{\text{fuse}}}$ are fed into the Video DiT $\epsilon_{\theta}$, trained with LoRA~\cite{hu2022lora} modules using diffusion loss~\cite{ho2020denoising, lipman2022flow, liu2022flow}:
\begin{equation}
    \mathcal{L}(\theta) = \mathbb{E}_{\bm{y}, \epsilon, \bm{f^{\text{fuse}}}, t} \big[\left\| \epsilon - \epsilon_{\theta}(\bm{y_t}, \bm{f^{\text{fuse}}}, t) \right\|_{2}^{2}\big],
\end{equation}
where $\bm{y_t}$ is the noisy video at diffusion timestep $t$, and $\epsilon$ is the random Gaussian noise.
In total, the loss for the Top-Down Memory Integration stage is formulated as:
\begin{equation}
\mathcal{L}_{\text {stage2}} =\mathcal{L}_{\text {stage1}} + \mathcal{L}(\theta).
\end{equation}

\subsection{Inference}
In contrast to the complex inference processes of previous methods~\cite{gong2024neuroclips, wang2025neurons}, which necessitate components like ControlNet~\cite{zhang2023adding}, keyframe reconstruction for the first frame of the video, and additional condition generation, our \textit{end-to-end} pipeline is straightforward and efficient. 
Users simply input fMRI signals into our Brain Model, and the Mixture-of-Memories then produces fused embeddings that condition the video decoder for seamless reconstruction.

\section{Experiment}

\subsection{Experimental Setup}
\label{sec:exp_setup}
\noindent\textbf{Datasets.}\quad
We conduct experiments on two publicly available fMRI-to-video datasets: cc2017~\cite{wen2018neural} and CineBrain~\cite{gao2025cinebrain}.  
The cc2017 dataset comprises fMRI data from three subjects who viewed various natural videos; the stimuli are split into 18 eight-minute training segments (2 repeats) and 5 eight-minute test segments (10 repeats), yielding around 11.5 hours of data per subject.
The CineBrain dataset provides synchronized EEG and fMRI recordings from six healthy participants as they watch and listen to 20 episodes of ``The Big Bang Theory'' (720p), totaling approximately 6 hours of audiovisual stimuli per subject. The fMRI data were sampled at 1.25 Hz.
We preprocess the cc2017 fMRI data following NeuroClips~\cite{gong2024neuroclips}.
For the CineBrain dataset, which provides visual ROIs, we follow the same pipeline to extract the additional hippocampus ROI for our experiments.
More details about preprocessing are provided in the Supplementary Materials~\ref{app:sec_preprocess}.

\noindent\textbf{Evaluation Metrics.}\quad
We evaluate the reconstructed videos at semantic, spatiotemporal, and pixel levels~\cite{gong2024neuroclips, gao2025cinebrain}. 
For semantic-level evaluation, we compute $N$-way top-$K$ accuracy to assess whether the generated videos semantically match the ground-truth (GT) clips, using a VideoMAE~\cite{tong2022videomae}-based classifier on 400 video classes from the Kinetics-400 dataset~\cite{kay2017kinetics}, following prior work~\cite{gong2024neuroclips, wang2025neurons, gao2025cinebrain}.
For spatiotemporal-level evaluation, we use 
CLIP temporal consistency (CLIP-pcc)~\cite{radford2021learning} and DINO~\cite{oquab2023dinov2} temporal consistency~\cite{huang2024vbench, gao2025cinebrain} (DTC) to measure the spatiotemporal coherence of the generated videos.
Additionally, we employ the Motion Smoothness (MS) and Dynamic Degree~\cite{huang2024vbench} to assess the smoothness and magnitude of movements, along with the Mean End-Point Error (EPE) to assess the motion consistency compared with GT videos.
Apart from video-based metrics above, we also assess frame-wise image quality at both semantic and pixel levels;
see Supplementary Materials~\ref{app:sec_metric} and~\ref{app:sec_frame_based_metrics} for more details and results. 

\noindent\textbf{Implementation Details.}\quad
We use the AdamW~\cite{loshchilov2017decoupled} optimizer with the OneCycle learning rate schedule~\cite{smith2019super}. 
In the first stage, the Brain Model is trained for 8,000 steps with a batch size of 144 and a learning rate of $1 \times 10^{-4}$.
In the second stage, the Brain Model 
along with the router and fusion module in Mixture-of-Memories are trained for 20 epochs with a batch size of 32 and a learning rate of $1 \times 10^{-6}$. Following~\cite{chen2023cinematic, gao2025cinebrain}, the transformer-based Brain Model consists of 24 layers, each with a hidden dimension of 2048 and a token length of 513 (512 fMRI embedding tokens plus one global fMRI token).
We use a LoRA rank of 16 with a scaling factor of 16.
We adopt Wan2.1 1.3B~\cite{wang2025wan} as our video decoder.
On the cc2017 dataset, we generate 57-frame videos at 624$\times$624, much longer and higher resolution than NeuroClips (16 frames at 256$\times$256) and MindVideo (6 frames at 256$\times$256).
On the CineBrain dataset, we generate 33 frames at 720$\times$480 resolution, matching the source videos.
More implementation details are provided in the Supplementary Materials~\ref{app:sec_implent_details}.

\subsection{Main Results}

\begin{figure*}[t]
  \centering
  \includegraphics[width=0.96\linewidth]{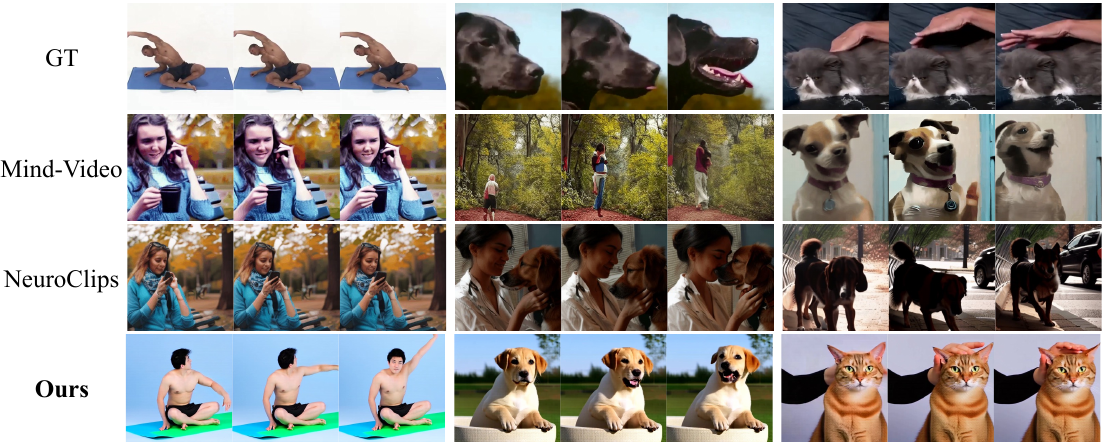}
  \caption{\textbf{Qualitative comparison of \frameworkplain and baselines on the cc2017 dataset}.}
  \label{fig:compare_wen}
\end{figure*}
\begin{table*}[t]
\caption{
\textbf{Quantitative comparison results of all subjects on the cc2017 and CineBrain datasets.}
Results for the cc2017 dataset are quoted from~\cite{gong2024neuroclips, lu2024animate}, except for the DTC and MS (Motion Smoothness) metrics. Results for the CineBrain dataset are quoted from~\cite{gao2025cinebrain}, except for the MS metric.
``*''~denotes methods reimplemented using the same decoder model and fMRI input as \frameworkplain.
}
\small
\centering
    \resizebox{\linewidth}{!}{
    \begin{tabular}{l|l|cc|ccc|cc}
    \toprule
    \multicolumn{1}{c|}{\multirow{2}{*}{\textsc{Dataset}}} & \multicolumn{1}{c|}{\multirow{2}{*}{\textsc{Methods}}} & \multicolumn{2}{c|}{Semantic-level} &
    \multicolumn{3}{c|}{Spatiotemporal-level} &
    \multicolumn{2}{c}{Pixel-level} \\
     & & 2-way & 50-way & CLIP-pcc & DTC & MS & SSIM & PSNR \\
    \midrule
    \midrule
\multirow{5}{*}{cc2017} & Kupershmidt~\cite{kupershmidt2022penny} & 0.771\scriptsize{$\pm0.03$} & - & 0.386\scriptsize{$\pm0.47$} & - & - & 0.135\scriptsize{$\pm0.08$} & 8.761\scriptsize{$\pm2.22$} \\
& MinD-Video~\cite{chen2023cinematic} & \underline{0.839\scriptsize{$\pm0.03$}} & {0.197\scriptsize{$\pm0.02$}} & {0.408\scriptsize{$\pm0.46$}} & 0.884\scriptsize{$\pm0.08$} & 0.901\scriptsize{$\pm0.05$} & 0.171\scriptsize{$\pm0.08$} & 8.662\scriptsize{$\pm1.52$} \\
& Mind-Animator~\cite{lu2024animate} & 0.830 & - & 0.425 & - & - & 0.321 & \underline{9.220} \\
& NeuroClips~\cite{gong2024neuroclips} & {0.834\scriptsize{$\pm0.03$}} & \underline{0.220\scriptsize{$\pm0.01$}} & \underline{0.738\scriptsize{$\pm0.17$}} & \underline{0.926\scriptsize{$\pm0.05$}} & \underline{0.955\scriptsize{$\pm0.01$}} & \textbf{0.390}\scriptsize{$\pm0.08$} & 9.211\scriptsize{$\pm1.46$} \\ 
& \cellcolor{cyan!10} \framework & \cellcolor{cyan!10} \textbf{0.850}\scriptsize{$\pm0.02$} & \cellcolor{cyan!10} \textbf{0.240}\scriptsize{$\pm0.03$} & \cellcolor{cyan!10} \textbf{0.972}\scriptsize{$\pm0.01$} & \cellcolor{cyan!10} \textbf{0.954}\scriptsize{$\pm0.01$} & \cellcolor{cyan!10} \textbf{0.966}\scriptsize{$\pm0.02$} & \cellcolor{cyan!10} \underline{0.375\scriptsize{$\pm0.06$}} & \cellcolor{cyan!10} \textbf{9.476\scriptsize{$\pm0.23$}}  \\
\midrule
\multirow{4}{*}{CineBrain} & GLFA~\cite{li2024enhancing} & 0.801 & 0.167 & 0.735 & 0.706 & - & 0.123 & 7.526 \\
& CineSync~\cite{gao2025cinebrain}  & 0.893  & 0.307 & 0.945 & 0.907 & \underline{0.974\scriptsize{$\pm0.01$}} & 0.240   & 11.92  \\ 
& CineSync*  & \underline{0.933\scriptsize{$\pm0.02$}}  & \underline{0.324\scriptsize{$\pm0.02$}}  & \underline{0.977\scriptsize{$\pm0.01$}} &  \underline{0.942\scriptsize{$\pm0.01$}} & 0.973\scriptsize{$\pm0.01$} & \underline{0.267}\scriptsize{$\pm0.04$} & \textbf{16.04}\scriptsize{$\pm2.35$}  \\ 
& \cellcolor{cyan!10} \framework & \cellcolor{cyan!10} \textbf{0.937}\scriptsize{$\pm0.02$} & \cellcolor{cyan!10} \textbf{0.393}\scriptsize{$\pm0.03$} & \cellcolor{cyan!10} \textbf{0.988}\scriptsize{$\pm0.01$} & \cellcolor{cyan!10} \textbf{0.975}\scriptsize{$\pm0.01$} & \cellcolor{cyan!10} \textbf{0.975}\scriptsize{$\pm0.01$} & \cellcolor{cyan!10} \textbf{0.271\scriptsize{$\pm0.05$}} & \cellcolor{cyan!10} \underline{16.02}\scriptsize{$\pm2.38$} \\
    \bottomrule
    \end{tabular}
    }
\label{tab:compare_all}
\vspace{-3mm}
\end{table*}

\noindent\textbf{Results on cc2017 Dataset.}\quad
Fig.~\ref{fig:compare_wen} reveals that the baselines suffer from semantic errors and low reconstruction quality. For example, in the left and right panels of Fig.~\ref{fig:compare_wen}, both MindVideo and NeuroClips fail to reconstruct the yoga and petting actions; in the middle panel, MindVideo incorrectly decodes the dog as a person, while NeuroClips hallucinates spurious objects (a woman) and background context (a room) that are absent from the GT video. In contrast, our \framework faithfully reconstructs complex videos with coherent action, category, and context semantics, demonstrating its superior performance.

Tab.~\ref{tab:compare_all} shows that our method achieves the best scores on semantic and spatiotemporal metrics, showing the effectiveness of our comprehensive semantic enrichment and the Mixture-of-Memories design. At the pixel level, we obtain the highest PSNR and competitive SSIM. Although NeuroClips reports slightly higher SSIM, its reconstructions exhibit inferior semantics and temporal coherence, indicating a trade-off between fidelity and semantic alignment. In contrast, our method strikes a better balance.
Note that Tab.~\ref{tab:compare_all} reports the average scores across all subjects; per-subject results are provided in Supplementary Materials~\ref{app:all_results}.

\begin{figure*}[t]
  \centering
  \includegraphics[width=0.96\linewidth]{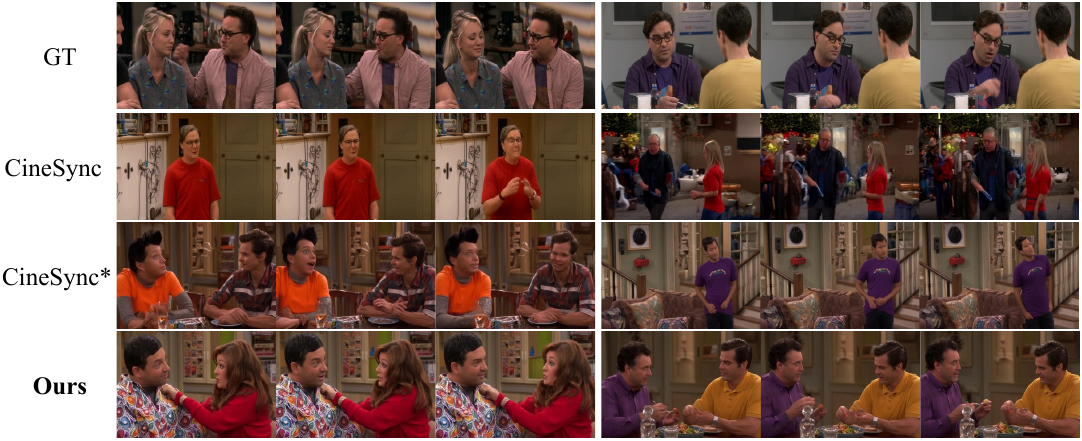}
  \caption{\textbf{Qualitative comparison of \frameworkplain and baselines on the CineBrain dataset.}}
  \label{fig:compare_cineBrain}
\end{figure*}
\begin{table*}[h!]
\caption{
\textbf{Quantitative comparison of motion consistency on the cc2017 and CineBrain datasets}. ``$\Delta$Dynamic Degree'' denotes the absolute difference from the GT videos' Dynamic Degree; lower values indicate motion amplitudes closer to the GT.
}
\label{tab:traj_metrics}
\centering
\small
\begin{tabular}{l|ccc|cc}
\toprule
 & \multicolumn{3}{c|}{\textbf{Test Set: cc2017}} & \multicolumn{2}{c}{\textbf{Test Set: CineBrain}} \\
\cmidrule(lr){2-4}\cmidrule(lr){5-6}
\textsc{Metric} & 
\textsc{MindVideo} & \textsc{NeuroClips} & \frameworkplain & \textsc{CineSync} & \frameworkplain \\
\midrule
EPE ($\downarrow$)            
& 9.045 & 4.432 & \textbf{1.628}
& 3.258 & \textbf{2.126} \\
$\Delta$Dynamic Degree ($\downarrow$)   
& 0.1155 & 0.2133 & \textbf{0.0200}
& 0.2611 & \textbf{0.0649} \\
\bottomrule
\end{tabular}
\vspace{-2mm}
\end{table*}

\noindent\textbf{Results on CineBrain Dataset.}\quad
Tab.~\ref{tab:compare_all} shows that our method outperforms all baselines across most metrics while achieving comparable PSNR with only marginal differences, confirming the broad effectiveness of our framework. 
Notably, our method also outperforms CineSync*, an enhanced baseline that uses the same decoder and additional hippocampal fMRI input as \frameworkplain.
Per-subject results are provided in Supplementary Materials~\ref{app:all_results}.

Fig.~\ref{fig:compare_cineBrain} shows that strong baselines (\eg, CineSync and CineSync*) still struggle to accurately reconstruct visual details or capture action semantics. In contrast, videos reconstructed from our method are temporally consistent and semantically accurate.

\noindent\textbf{Comparison on Motion Consistency.}\quad
Tab.~\ref{tab:traj_metrics} shows that our method achieves the lowest (best) endpoint error (EPE) on both cc2017 and CineBrain datasets, indicating the closest match to GT motion trajectories.
We further assess motion magnitude using the Dynamic Degree metric~\cite{huang2024vbench}, reporting the absolute difference from the GT (lower is better). Our method achieves the smallest discrepancy, closely matching the GT motion magnitude.

\noindent\textbf{Human Evaluation Results.}\quad
We conduct a comprehensive human evaluation to further evaluate the video reconstruction performance. Twenty participants evaluate 360 video groups, comparing videos from four anonymous methods (MindVideo, NeuroClips, Mind-Animator, and our \frameworkplain) to the GT video over four dimensions: 1) Semantic Alignment, 2) Temporal Consistency, 3) Visual Quality, and 4) Overall Fidelity. The results in Tab.~\ref{tab:human_evaluation} show that \frameworkplain significantly outperforms competing methods across all dimensions.

\begin{table}[t]
\caption{
\textbf{Human evaluation results.}
}
\small
\centering
    \resizebox{\columnwidth}{!}{
    \begin{tabular}{l|rrrr}
    \toprule
    \textsc{Method} & \tabincell{c}{Semantic \\ Alignment} & \tabincell{c}{Temporal \\ Consistency} & \tabincell{c}{Visual \\ Quality} & \tabincell{c}{Overall \\ Fidelity} \\
    \midrule
    MindVideo	    & 8.31\%	& 7.74\%	& 9.08\%	& 7.79\% \\
    NeuroClips	    & 16.30\%	& 18.13\%	& 13.79\%	& 15.80\% \\
    Mind-Animator	& 11.62\%	& 8.23\%	& 6.46\%	& 8.82\% \\
    \rowcolor{cyan!10}
    Ours	        & \textbf{63.77\%} & \textbf{65.90\%} & \textbf{70.67\%} & \textbf{67.59\%} \\  
    \bottomrule
    \end{tabular}
    }
\label{tab:human_evaluation}
\vspace{-4mm}
\end{table}

\subsection{Ablation Studies}
We conduct ablation studies to evaluate each component's effectiveness on cc2017 dataset, and the impact of using additional hippocampus fMRI data on CineBrain dataset.

\begin{figure}
    \centering
    \includegraphics[width=1.0\linewidth]{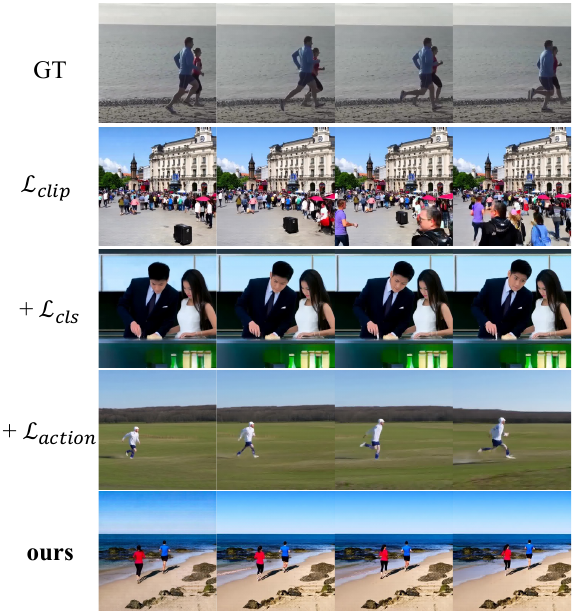}
    \caption{\textbf{Qualitative ablation of each proposed component.}}
    \label{fig:qualitative_ablation}
    \vspace{-2mm}
\end{figure}

\noindent\textbf{Ablation on Each Proposed Component.}\quad
Quantitative (Tab.~\ref{tab:ablation}) and qualitative (Fig.~\ref{fig:qualitative_ablation}) ablation results highlight the effectiveness of each component.
For ablation baselines without Mixture-of-Memories (MoM), we condition the video decoder on the retrieved most similar (top-$1$) text embedding for reconstruction.
We observe that the baseline using only $\mathcal{L}_{\text{clip}}$ struggles to generate semantically accurate videos. 
Introducing the classification loss $\mathcal{L}_{\text{cls}}$ improves object recognition (\eg, identifying the object as two persons, as in Fig.~\ref{fig:qualitative_ablation}), but still fails to capture correct actions. Adding the action contrastive loss $\mathcal{L}_{\text{action}}$ further helps the model grasp action concepts (\eg, detecting that a person is running), though details remain inaccurate. Finally, incorporating the MoM module allows the model to accurately perceive both object categories and actions, and effectively reconstruct fine-grained details.

\begin{figure*}[t]
  \centering
  \includegraphics[width=1.0\linewidth]{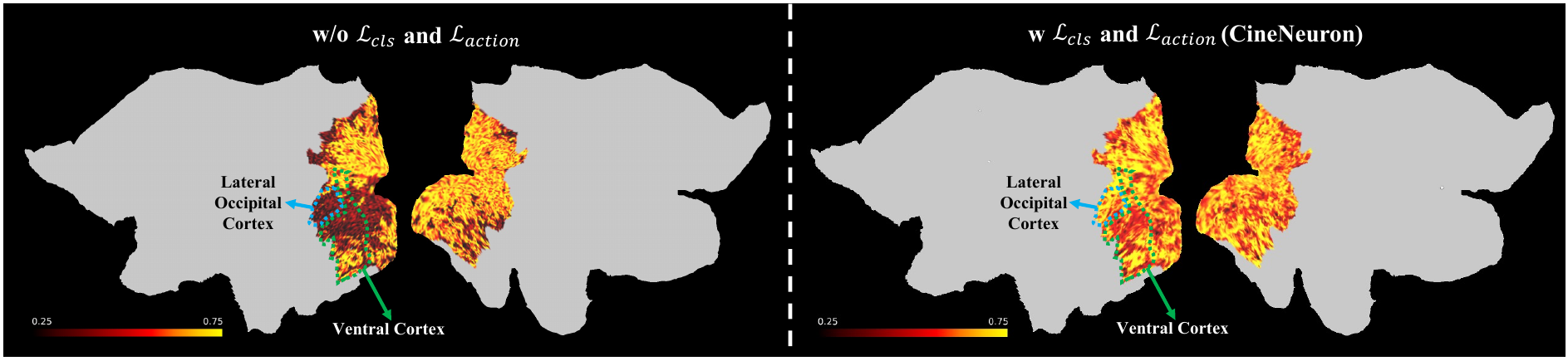}
  \caption{\textbf{Visualization of voxel weights from the first regression layer.} 
  Voxel weights are averaged and normalized to $[0, 1]$, displayed with a 0.25 to 0.75 colorbar.
  The blue and green dotted lines indicate the lateral occipital and ventral cortex, respectively.}
  \label{fig:inter}
\end{figure*}
\begin{table}[t]
\caption{
\label{tab:ablation}
\textbf{Quantitative ablation of each component on Subject 1 of the cc2017 dataset.}
}
\small
\centering
    \resizebox{\linewidth}{!}{
    \begin{tabular}{cccc|cc|ccc}
    \toprule
    \multicolumn{4}{c|}{\multirow{1}{*}{\textsc{Key Components}}} & \multicolumn{2}{c|}{Semantic-level} & \multicolumn{3}{c}{Spatiotemporal-level} \\
     $\mathcal{L}_{\text{clip}}$ & $\mathcal{L}_{\text{cls}}$ & 
     $\mathcal{L}_{\text{action}}$
     & MoM
     & 2-way & 50-way & CLIP-pcc & DTC & MS \\
    \midrule
    \midrule
\cmark & &  &  & 0.824\scriptsize{$\pm$0.01} & 0.217\scriptsize{$\pm$0.03} & \textbf{0.973}\scriptsize{$\pm$0.01}  & \textbf{0.960}\scriptsize{$\pm$0.03} & 0.961\scriptsize{$\pm$0.01} \\ 
\cmark & \cmark & & & 0.829\scriptsize{$\pm$0.01} & \underline{0.228\scriptsize{$\pm$0.03}} & 0.965\scriptsize{$\pm$0.01} & 0.951\scriptsize{$\pm$0.01} & 0.956\scriptsize{$\pm$0.02} \\ 
\cmark & \cmark & \cmark & & \underline{0.835\scriptsize{$\pm$0.02}} & 0.223\scriptsize{$\pm$0.04} & \underline{0.970\scriptsize{$\pm$0.01}} & 0.957\scriptsize{$\pm$0.01} & \underline{0.963\scriptsize{$\pm$0.01}} \\ 
\midrule
\rowcolor{cyan!10} \cmark & \cmark & \cmark & \cmark & \textbf{0.846}\scriptsize{$\pm$0.02}  & \textbf{0.237}\scriptsize{$\pm$0.02} & \textbf{0.973}\scriptsize{$\pm$0.01} & \underline{0.959\scriptsize{$\pm$0.01}} & \textbf{0.967}\scriptsize{$\pm$0.02} \\
    \bottomrule
    \end{tabular}
    }
\vspace{-4mm}
\end{table}

\noindent\textbf{Ablation on the Input fMRI Data.}\quad
By default, on the CineBrain dataset, \frameworkplain takes fMRI data from both the visual area and hippocampus as input for video reconstruction. 
As shown in the second row of Tab.~\ref{tab:ablation_fMRI}, removing the hippocampus fMRI input causes a performance drop in semantic-level metrics while spatiotemporal metrics remain comparable. This indicates that the hippocampus provides crucial semantic information, validating our top-down memory integration design.
Furthermore, we test the effect of adding fMRI signals from the mPFC region, which is known to collaborate with the hippocampus in memory processing. As shown in the third row of Tab.~\ref{tab:ablation_fMRI}, including mPFC inputs yields further performance gains, especially on semantic metrics. This result reinforces our model's design, which integrates retrieved memories (long-term) with current fMRI features (working memory).

\noindent\textbf{Ablation on Integration Step in MoM.}
Tab.~\ref{tab:ablation_brainfuser} ablates the MoM integration step and shows it is the main contributor to spatiotemporal quality gains. Prior work like MindVideo directly conditions the generator on raw fMRI features, which are misaligned with the pretrained model's text-anchored conditioning space, leading to poor spatiotemporal quality. In contrast, our integration step in MoM uses residual fusion with retrieved multimodal memories to supplement details missing from fMRI while preserving the stability of the generator's feature space, producing higher-quality videos.

\begin{table}[t]
\caption{
\textbf{Quantitative ablation of hippocampus and mPFC fMRI inputs on Subject 5 of the CineBrain dataset.}
}
\centering
    \resizebox{\columnwidth}{!}{
    \begin{tabular}{l|cc|ccc}
    \toprule
    \multicolumn{1}{c|}{\multirow{2}{*}{\textsc{Methods}}} & \multicolumn{2}{c|}{Semantic-level} &
    \multicolumn{3}{c}{Spatiotemporal-level} \\
     & 2-way & 50-way & CLIP-pcc & DTC & MS \\
    \midrule
\cellcolor{cyan!10}\framework & \cellcolor{cyan!10}0.938\scriptsize{$\pm$0.01} & \cellcolor{cyan!10}0.401\scriptsize{$\pm$0.02} & \cellcolor{cyan!10}0.990\scriptsize{$\pm$0.01} & \cellcolor{cyan!10}0.978\scriptsize{$\pm$0.01} & \cellcolor{cyan!10}0.976\scriptsize{$\pm$0.01} \\  
 -- hippo. input & 0.935\scriptsize{$\pm$0.02} & 0.358\scriptsize{$\pm$0.03} & 0.993\scriptsize{$\pm$0.01} & 0.974\scriptsize{$\pm$0.01} & 0.975\scriptsize{$\pm$0.01} \\ 
+ mPFC & 0.945\scriptsize{$\pm$0.01} & 0.442\scriptsize{$\pm$0.03} & 0.993\scriptsize{$\pm$0.01} & 0.981\scriptsize{$\pm$0.01} & 0.976\scriptsize{$\pm$0.01} \\
    \bottomrule
    \end{tabular}
    }
\label{tab:ablation_fMRI}
\end{table}

\begin{table}[t]
\caption{\textbf{Quantitative ablation of MoM's integration step on Subject 1 of the cc2017 dataset.}}
\label{tab:ablation_brainfuser}
\centering
\small
    \resizebox{\columnwidth}{!}{
    \begin{tabular}{l|cc|ccc}
    \toprule
    \multicolumn{1}{c|}{\multirow{2}{*}{\textsc{Methods}}} & \multicolumn{2}{c|}{Semantic-level} & \multicolumn{3}{c}{Spatiotemporal-level} \\
    \cmidrule(lr){2-3} \cmidrule(lr){4-6}
    & 2-way & 50-way & CLIP-pcc & DTC & MS \\
    \midrule
    w/o Integration & 0.830\scriptsize{$\pm$0.02} & 0.224\scriptsize{$\pm$0.05} & 0.802\scriptsize{$\pm$0.01} & 0.920\scriptsize{$\pm$0.01} & 0.961\scriptsize{$\pm$0.01} \\
    \rowcolor{cyan!10}
    w/ Integration & \textbf{0.846\scriptsize{$\pm$0.02}} & \textbf{0.237\scriptsize{$\pm$0.02}} & \textbf{0.973\scriptsize{$\pm$0.01}} & \textbf{0.959\scriptsize{$\pm$0.01}} & \textbf{0.967\scriptsize{$\pm$0.02}} \\
    \bottomrule
    \end{tabular}
    }
    \vspace{-4mm}
\end{table}

\subsection{Interpretation Results}
To assess the neural interpretability of our model, we visualize voxel weights within the visual cortex on a brain flat map, comparing models trained without and with the combination of category semantic learning $\mathcal{L}_{\text{cls}}$ and action semantic alignment $\mathcal{L}_{\text{action}}$ on the CineBrain dataset, as shown in Fig.~\ref{fig:inter}. 
Incorporating these tasks significantly increases voxel weights in the lateral occipital and ventral visual cortices---regions that are critically related to object recognition, shape analysis, and complex visual processing~\cite{kourtzi2001representation, grill2014functional}.
Notably, weight increases also appear in visual motion-sensitive areas (\eg~V5/MT, MST, and FST) within the lateral occipital cortex, highlighting the effectiveness of the designed tasks in enhancing both object identification and visual motion capture~\cite{ffytche1995parallel, born2005structure}.

\section{Conclusion}
In this paper, we present \frameworkplain, a hierarchical framework for semantically enhanced fMRI-to-video reconstruction that combines bottom-up semantic enrichment with top-down memory integration. We enrich fMRI embeddings with comprehensive video semantics from text, image, action, and category modalities. Building on these embeddings, our Mixture-of-Memories adaptively retrieves multimodal memories and fuses them into the fMRI representation to refine video reconstruction.
Extensive experiments on two datasets show that \frameworkplain surpasses prior methods on both qualitative and quantitative results. Furthermore, we provide an interpretable analysis demonstrating functional alignment between the human visual cortex and our model, offering insights into neural processing.

\newpage
\noindent\textbf{Acknowledgements.}\quad This work was supported in part by National Natural Science Foundation of China (Nos. T2596013 and 62471148), STI2030-Major Projects (No. 2021ZD0200204), and Shanghai Center for Brain Science and Brain-inspired Technology.
{
    \small
    \bibliographystyle{ieeenat_fullname}
    \bibliography{main}
}

\clearpage
\setcounter{page}{1}

\appendix
\onecolumn

\begin{center}
	\Large \textbf{Bridging Brain and Semantics: A Hierarchical Framework for \\ Semantically Enhanced fMRI-to-Video Reconstruction} \\
    Supplementary Material
\end{center}

\phantomsection\label{toc:appendix}

\etocdepthtag.toc{mtappendix}
\etocsettagdepth{mtchapter}{none}
\etocsettagdepth{mtappendix}{subsection}
\tableofcontents

\clearpage
\twocolumn

\setcounter{table}{0}
\setcounter{figure}{0}
\setcounter{equation}{0}
\renewcommand{\thetable}{S\arabic{table}}
\renewcommand{\thefigure}{S\arabic{figure}}
\renewcommand{\theequation}{S\arabic{equation}}

In this supplementary material, we provide comprehensive details and additional experimental results to support the findings in the main text. First, Sec.~\ref{app:exp_setup} elaborates on the experimental setup, including data preprocessing pipelines, detailed metric definitions, and implementation specifics. Sec.~\ref{app:cls_task_construction} describes the construction of the category semantic learning task. Sec.~\ref{app:more_exp_results} presents extensive quantitative and qualitative evaluation of \frameworkplain. Sec.~\ref{app:more_ablation} presents further ablation studies on the EEG input, superclass processing, and different semantics incorporated in \frameworkplain. Finally, Sec.~\ref{sec:limitation} and Sec.~\ref{app:impacts} discuss the limitations and ethical considerations of our work, respectively.

\section{Experimental Setup}
\label{app:exp_setup}
In this section, we provide a comprehensive description of the experimental setup. We first detail the data preprocessing pipelines and voxel selection procedures for the cc2017 and CineBrain datasets in Sec.~\ref{app:sec_preprocess}. Next, we define the frame-based evaluation metrics used to assess semantic and pixel-level quality in Sec.~\ref{app:sec_metric}. Finally, we present additional implementation details and specific hyperparameter settings in Sec.~\ref{app:sec_implent_details}.

\subsection{Details of Datasets and Data Preprocessing}
\label{app:sec_preprocess}

\paragraph{cc2017 Dataset.}
We preprocess fMRI data in the cc2017 dataset using the pipelines described in~\cite{gong2024neuroclips, wang2025neurons}, following the procedures outlined in~\cite{glasser2013minimal}. The pipeline consists of five stages: artifact removal, motion correction using six degrees of freedom, registration to MNI standard space, transformation to cortical surfaces, and subsequent coregistration to a cortical surface template as detailed in~\cite{glasser2016multi}. 
In line with~\cite{gong2024neuroclips}, we select stimulus-activated voxels by evaluating voxel-wise correlations of training videos. These correlations are processed through Fisher z-transformation and assessed using a one-sample t-test. 
We then select voxels with significant activation (Bonferroni-corrected, $P < 0.05$), resulting in 13,447, 14,828, and 9,114 activated voxels in the visual cortex for Subjects 1, 2, and 3, respectively.
The training videos are processed into 2-second clips to match the 2-second temporal resolution of the fMRI data. These clips have a resolution of $57 \times 624 \times 624$ to align with the input requirements of the video decoder model (Wan2.1).
As a result, we obtain 8,640 (4,320$\times$2) fMRI-video pairs for training and 1,200 pairs for testing.
Additionally, we incorporate a 4-second delay in the BOLD signals to account for hemodynamic response latency when mapping movie stimulus responses, as suggested by~\cite{han2019variational, nishimoto2011reconstructing, wang2022reconstructing}.

\paragraph{CineBrain Dataset.}
We preprocess fMRI data from the CineBrain dataset utilizing the processes described in~\cite{gao2025cinebrain}, specifically employing the widely adopted fMRIPrep pipeline~\cite{esteban2019fmriprep}. The fMRI data are collected at a frequency of 1.25 Hz, corresponding to a temporal resolution of 0.8 seconds. 
The selected visual regions of interest (ROIs) in the CineBrain dataset are characterized using the parcellation provided by the Human Connectome Project Multi-Modal Parcellation (HCP-MMP) within the 32k\_fs\_LR space.
The identified visual ROIs include areas such as ``V1, V2, V3, V3A, V3B, V3CD, V4, LO1, LO2, LO3, PIT, V4t, V6, V6A, V7, V8, PH, FFC, IP0, MT, MST, FST, VVC, VMV1, VMV2, VMV3, PHA1, PHA2, PHA3'', totaling 8,405 voxels in the visual cortex for each subject.
Additionally, we select 1,559 voxels within the hippocampus using the same pipeline applied to the visual regions.
Therefore, unless stated otherwise, our experiments on the CineBrain dataset utilize a total of 9,964 voxels from the visual cortex and hippocampus as fMRI data.
The videos viewed by participants are standardized to 18 minutes and segmented into 4-second clips to align with the fMRI temporal resolution of 5 fMRI signals, each with a resolution of 0.8 seconds, as described in CineBrain.
Consequently, we obtain 4,860 training samples and 540 testing samples of fMRI-video pairs with resolution $33\times480\times720$.
In addition, fMRI signals underwent z-scoring across vertices to adjust for the inherent delay in BOLD responses, factoring in a 4-second lag.

\subsection{Details of Frame-Based Metrics}
\label{app:sec_metric}
In addition to the video-level metrics described in the main text, 
following~\cite{gong2024neuroclips, wang2025neurons, gao2025cinebrain}, we evaluate the generated frames at both semantic and pixel levels.
For semantic-level evaluation, we perform an $N$-way top-$K$ accuracy classification test on 1,000 ImageNet~\cite{deng2009imagenet} classes using an ImageNet classifier.
A trial is successful if the ground truth (GT) class is among the top-$1$ probabilities in the generated frame results, selected from $N$ random classes, including the GT class.
For pixel-level evaluation, we employ the Structural Similarity Index (SSIM)~\cite{wang2004ssim} and Peak Signal-to-Noise Ratio (PSNR) to assess the image quality of video frames.

\subsection{Implementation Details}
\label{app:sec_implent_details}
In addition to Sec.~4.1, we provide more implementation details here. The temperature $\tau$ is set to 0.07. The loss weights $\lambda_1$ and $\lambda_2$ are set to 0.1 and 10, respectively, while $\alpha$ in~Eq.~(\ref{eq:fuse_embed}) is set to 1. 
The routing network within the Mixture-of-Memories (MoM) is implemented as a linear layer. 
We train our model and conduct experiments on 4 A800 GPUs.

\begin{figure*}[t]
  \centering
  \includegraphics[width=1.0\linewidth]{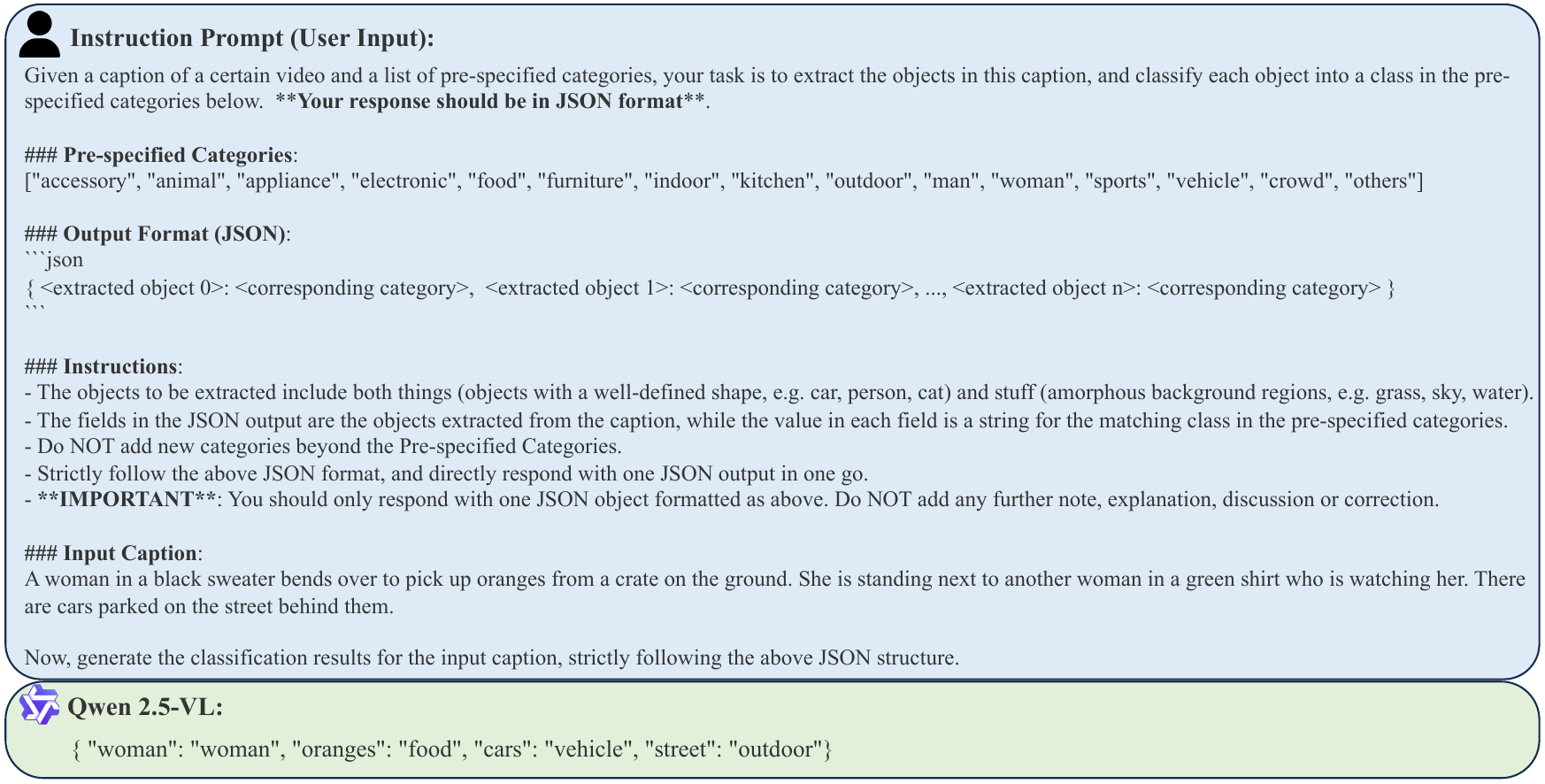}
  \caption{The instruction prompts used for constructing the object category recognition task.}
  \label{fig:app_class_task}
\end{figure*}
\begin{table*}[h]
\caption{
\textbf{Quantitative results of all subjects on the cc2017 and CineBrain datasets.} Please note that, currently, the CineBrain dataset provides fMRI data only for subjects 1 and 5.}
\label{tab:app_all_subjects}
\small
\centering
    \begin{tabular}{l|l|cc|ccc}
    \toprule
    \multicolumn{1}{c|}{\multirow{2}{*}{\textsc{Dataset}}} & \multicolumn{1}{c|}{\multirow{2}{*}{\textsc{Methods}}} & \multicolumn{2}{c|}{Semantic-level} &
    \multicolumn{3}{c}{Spatiotemporal-level} \\
     & & 2-way & 50-way & CLIP-pcc & DTC & MS \\
    \midrule
    \midrule
\multirow{4}{*}{cc2017} & Subject1 &  0.846\scriptsize{$\pm0.02$}  &  0.237\scriptsize{$\pm0.02$} &  0.973\scriptsize{$\pm0.01$} &  0.959\scriptsize{$\pm0.01$} &  0.967\scriptsize{$\pm0.02$} \\ 
 & Subject2 & 0.852\scriptsize{$\pm0.02$} & 0.241\scriptsize{$\pm0.04$} &  0.972\scriptsize{$\pm0.01$} &  0.956\scriptsize{$\pm0.01$} &  0.966\scriptsize{$\pm0.01$}\\ 
 & Subject3 &  0.853\scriptsize{$\pm0.02$}  &  0.242\scriptsize{$\pm0.04$} &  0.971\scriptsize{$\pm0.01$} &  0.948\scriptsize{$\pm0.01$} &  0.964\scriptsize{$\pm0.02$}\\ 
 & \cellcolor{cyan!10} Average & \cellcolor{cyan!10} 0.850\scriptsize{$\pm0.02$} & \cellcolor{cyan!10} 0.240\scriptsize{$\pm0.03$} & \cellcolor{cyan!10} 0.972\scriptsize{$\pm0.01$} & \cellcolor{cyan!10} 0.954\scriptsize{$\pm0.01$} & \cellcolor{cyan!10} 0.966\scriptsize{$\pm0.02$} \\ 
\midrule
\multirow{3}{*}{CineBrain} & Subject1 & 0.936\scriptsize{$\pm0.02$} &  0.384\scriptsize{$\pm0.03$} &  0.986\scriptsize{$\pm0.01$} &  0.971\scriptsize{$\pm0.01$} &  0.974\scriptsize{$\pm0.01$} \\ 
 & Subject5 &  0.938\scriptsize{$\pm0.01$} &  0.401\scriptsize{$\pm0.02$} &  0.990\scriptsize{$\pm0.01$} &  0.978\scriptsize{$\pm0.01$} &  0.976\scriptsize{$\pm0.01$} \\ 
 & \cellcolor{cyan!10} Average & \cellcolor{cyan!10} 0.937\scriptsize{$\pm0.02$} & \cellcolor{cyan!10} 0.393\scriptsize{$\pm0.03$} & \cellcolor{cyan!10} 0.988\scriptsize{$\pm0.01$} & \cellcolor{cyan!10} 0.975\scriptsize{$\pm0.01$} & \cellcolor{cyan!10} 0.975\scriptsize{$\pm0.01$} \\  
    \bottomrule
    \end{tabular}
\end{table*}

\section{Classification Task Construction}
\label{app:cls_task_construction}
This section outlines the construction of the category semantic learning task used to enrich fMRI representations. We describe the automated pipeline for extracting object nouns from video captions using Qwen2.5-VL and mapping them to a simplified set of superclasses to facilitate robust classification and address class imbalance.

Specifically, we generate category names for the key objects in each video clip to establish the object category recognition task. To streamline this process, we directly utilize video captions generated by Qwen2.5-VL to extract object nouns (\eg, [`woman', `car', `oranges', ...]).
We then use Qwen2.5-VL to classify these nouns into the pre-defined MSCOCO~\cite{lin2014microsoft} categories, and simplify the categories by filtering and merging infrequent classes into a reduced set of 15 superclasses for both the cc2017 and CineBrain datasets.
The output is formatted as a JSON object containing the names of superclasses.
Detailed instruction prompts for category name generation, along with the superclass lists, are presented in Fig.~\ref{fig:app_class_task}.

\begin{table*}[!t]
\caption{
\textbf{Quantitative comparison results of frame-based metrics on the CineBrain dataset.} Results for the CineBrain dataset are quoted from~\cite{gao2025cinebrain}.
``*''~denotes methods reimplemented using the same decoder model and fMRI input as \frameworkplain.
}
\label{tab:app_compare_frame}
\small
\centering
    \begin{tabular}{l|cc|cc}
    \toprule
    \multicolumn{1}{c|}{\multirow{2}{*}{\textsc{Methods}}} & \multicolumn{2}{c|}{Semantic-level} &
    \multicolumn{2}{c}{Pixel-level} \\
     & 2-way & 50-way & SSIM & PSNR \\
    \midrule
    \midrule
GLFA~\cite{li2024enhancing} & 0.847 & 0.225 & 0.123 &  7.526  \\ 
CineSync~\cite{gao2025cinebrain}  & \underline{0.926} & \underline{0.358} & 0.240   & 11.92 \\ 
CineSync*  &  \underline{0.926}\scriptsize{$\pm0.04$} & 0.293\scriptsize{$\pm0.03$} & \underline{0.267}\scriptsize{$\pm0.04$} & \textbf{16.04}\scriptsize{$\pm2.35$} \\ 
\midrule
\cellcolor{cyan!10} \framework (Avg) & \cellcolor{cyan!10} \textbf{0.949\scriptsize{$\pm0.03$}} & \cellcolor{cyan!10} \textbf{0.438\scriptsize{$\pm0.04$}} & \cellcolor{cyan!10} \textbf{0.271\scriptsize{$\pm0.05$}} & \cellcolor{cyan!10} \underline{16.02}\scriptsize{$\pm2.38$} \\ 
Subject1 & 0.946\scriptsize{$\pm0.02$} & 0.425\scriptsize{$\pm0.05$} & 0.272\scriptsize{$\pm0.05$} & 16.11\scriptsize{$\pm2.65$} \\
Subject5 & 0.951\scriptsize{$\pm0.02$} & 0.451\scriptsize{$\pm0.03$} & 0.269\scriptsize{$\pm0.04$} & 15.92\scriptsize{$\pm2.11$} \\
    \bottomrule
    \end{tabular}
\end{table*}

\section{More Experimental Results}\label{app:more_exp_results}

In this section, we provide a comprehensive evaluation of our \framework through various quantitative and qualitative analyses. We first present the fMRI-to-video reconstruction performance for each individual subject (Sec.~\ref{app:all_results}) and evaluate the model using frame-based semantic and pixel-level metrics (Sec.~\ref{app:sec_frame_based_metrics}). To demonstrate the versatility of our approach, we show its extension to alternative backbones like MindEye and NeuroClips (Sec.~\ref{app:extension_to_other_arch}). We further validate the scalability of our Mixture-of-Memories (MoM) strategy by expanding the memory pool with external datasets (Sec.~\ref{app:expand_memory_pool}). Additionally, we provide qualitative visualizations of reconstructed videos (Sec.~\ref{app:more_visualization}) and assess the model's robustness regarding out-of-distribution (OOD) concepts (Sec.~\ref{app:ood}), cross-dataset generalization on BOLDMoments (Sec.~\ref{app:boldmoments}), and neural interpretability (Sec.~\ref{app:more_interpretation}). 
We further perform sensitivity analysis of the model's robustness to noise (Sec.~\ref{app:ablation_noise}). Finally, we report detailed metrics for retrieval accuracy (Sec.~\ref{app:retrieval_acc}) and classification performance (Sec.~\ref{app:cls_acc}).

\subsection{Results of Each Subject}
\label{app:all_results}
We report the fMRI-to-video reconstruction performance of our \framework on each subject in the cc2017 and CineBrain datasets in Tab.~\ref{tab:app_all_subjects}. 
The cc2017 dataset contains fMRI data from 3 subjects, while the CineBrain dataset currently provides fMRI data only for Subjects 1 and 5.
The results in Tab.~\ref{tab:app_all_subjects} demonstrate that our method consistently outperforms the baselines in average performance across all subjects, validating its effectiveness and robustness.

\subsection{Results of Frame-Based Metrics}
\label{app:sec_frame_based_metrics}
We provide frame-based evaluation results in Tab.~\ref{tab:app_compare_frame}, along with average results for our method across all subjects.
The results show that \framework excels in semantic-level frame understanding while maintaining competitive pixel-level performance.
In Tab.~\ref{tab:app_compare_frame}, \framework achieves the highest semantic-level 2-way accuracy, exceeding CineSync and CineSync* by 2.3\% and GLFA by 10.2\%. For 50-way semantic classification, our method provides an 8\% improvement over CineSync, demonstrating its ability to recognize fine-grained semantics.
Regarding pixel-level metrics, \framework achieves the best SSIM, surpassing CineSync by 3.1\%, and performs second-best in PSNR, competitive with CineSync*.
These results demonstrate that our method improves frame-based video comprehension by effectively integrating multimodal semantics.

\begin{table*}[!t]
\caption{
\textbf{Quantitative results of our \framework based on MindEye/NeuroClips architecture (denoted as $\dag$) on Subject 1 of cc2017 dataset.}
}
\small
\centering
    \begin{tabular}{l|cc|ccc}
    \toprule
    \multicolumn{1}{c|}{\multirow{2}{*}{\textsc{Methods}}} & \multicolumn{2}{c|}{Semantic-level} &
    \multicolumn{3}{c}{Spatiotemporal-level} \\
     & 2-way & 50-way & CLIP-pcc & DTC & MS \\
    \midrule
    \midrule
Wen~\cite{wen2018neural} & - & 0.166\scriptsize{$\pm0.02$} & - & - & - \\
Wang~\cite{wang2022reconstructing} & 0.773\scriptsize{$\pm0.03$} & - & 0.402\scriptsize{$\pm0.41$} & - & - \\
Kupershmidt~\cite{kupershmidt2022penny} & 0.771\scriptsize{$\pm0.03$} & - & 0.386\scriptsize{$\pm0.47$} & - & - \\
MinD-Video~\cite{chen2023cinematic} & 0.839\scriptsize{$\pm0.03$} & {0.197\scriptsize{$\pm0.02$}} & {0.408\scriptsize{$\pm0.46$}} & 0.884\scriptsize{$\pm0.08$} & 0.901\scriptsize{$\pm0.05$} \\
NeuroClips~\cite{gong2024neuroclips} & {0.834\scriptsize{$\pm0.03$}} & 0.220\scriptsize{$\pm0.01$} & 0.738\scriptsize{$\pm0.17$} & 0.926\scriptsize{$\pm0.05$} & 0.955\scriptsize{$\pm0.01$} \\ 
\midrule
 \framework &  \underline{0.846\scriptsize{$\pm0.02$}}  &  \underline{0.237\scriptsize{$\pm0.02$}} &  \underline{0.973\scriptsize{$\pm0.01$}} &  \textbf{0.959\scriptsize{$\pm0.01$}} &  \textbf{0.967\scriptsize{$\pm0.02$}} \\
 \rowcolor{cyan!10} \frameworkNoSpace$^\dag$ & \textbf{0.860\scriptsize{$\pm0.02$}}  &  \textbf{0.242\scriptsize{$\pm0.02$}} &  \textbf{0.974\scriptsize{$\pm0.01$}} &  \underline{0.958\scriptsize{$\pm0.01$}} &  \underline{0.965\scriptsize{$\pm0.01$}} \\
    \bottomrule
    \end{tabular}
\label{tab:app_extension}
\end{table*}
\begin{table*}[!t]
\caption{
\textbf{Quantitative ablation study of expanding the memory pool on Subject 1 of the cc2017 dataset.}
}
\small
\centering
\begin{tabular}{c|cc|cc|cc}
\toprule
\multicolumn{1}{c|}{\multirow{2}{*}{\textsc{Pool Size}}} & \multicolumn{2}{c|}{Semantic-level} & \multicolumn{2}{c|}{Pixel-level} & \multicolumn{2}{c}{Spatiotemporal-level} \\
& \textbf{Acc$_2$} & \textbf{Acc$_{50}$} & \textbf{SSIM} & \textbf{PSNR} & \textbf{CLIP-pcc} & \textbf{DTC} \\
\midrule
\midrule
Reduce 50\%(2160) & 0.845 & 0.235 & 0.372 & 9.429 & 0.970 & 0.949 \\
Ours (4320) & 0.846 & 0.237 & \textbf{0.376} & \underline{9.474} & \underline{0.973} & \underline{0.959} \\
\quad+AnimalKindom (5185) & \underline{0.854} & \underline{0.243} & 0.373 & 9.434 & \textbf{0.976} & \textbf{0.964} \\
\quad\quad+BOLDMoments (6185) & \textbf{0.858} & \textbf{0.254} & \underline{0.374} & \textbf{9.486} & 0.970 & \underline{0.959} \\
\bottomrule
\end{tabular}
\label{tab:app_ablation_expand_pool}
\end{table*}

\begin{table*}[!t]
\caption{
\textbf{Comparison on OOD samples of the cc2017 dataset.}
}
\small
\centering
\begin{tabular}{ccccccc}
\toprule
\textsc{Method} & \textbf{Acc$_2$} & \textbf{Acc$_{50}$} & \textbf{SSIM} & \textbf{PSNR} & \textbf{DTC} & \textbf{CLIP-pcc} \\ 
\midrule
\midrule
NeuroClips & \underline{0.801} & \underline{0.193} & 0.210 & \underline{9.193} & \underline{0.926} & \underline{0.959} \\
MindAnimator & 0.784 & 0.162 & \textbf{0.274} & 9.082 & 0.606 & 0.829 \\
\rowcolor{cyan!10} \textbf{Ours} & \textbf{0.821} & \textbf{0.197} & \underline{0.267} & \textbf{9.237} & \textbf{0.953} & \textbf{0.970} \\ 
\bottomrule
\end{tabular}
\label{tab:re_comparison_ood}
\end{table*}

\begin{table*}[!t]
\caption{
\textbf{Comparison on BOLDMoments~\cite{lahner2023bold} dataset.}
}
\small
\centering
\begin{tabular}{ccccccc}
\toprule
\textsc{Method} & \textbf{Acc$_2$} & \textbf{Acc$_{50}$} & \textbf{SSIM} & \textbf{PSNR} & \textbf{DTC} & \textbf{CLIP-pcc} \\ 
\midrule
\midrule
NeuroClips & 0.736 & 0.154 & 0.181 & 8.997 & 0.922 & 0.973 \\
\rowcolor{cyan!10} \textbf{Ours} & \textbf{0.791} & \textbf{0.192} & \textbf{0.230} & \textbf{9.078} & \textbf{0.968} & \textbf{0.986} \\ 
\bottomrule
\end{tabular}
\label{tab:re_comparison_bold}
\end{table*}

\subsection{Extension to Other Architectures}
\label{app:extension_to_other_arch}
We highlight that our \framework is architecture-agnostic and can extend to other architectures.
In addition to transformer-based architectures such as MindVideo~\cite{chen2023cinematic}, our method is also applicable to the architectures of MindEye~\cite{scotti2023reconstructing} and NeuroClips~\cite{gong2024neuroclips}.

MindEye and NeuroClips employ an MLP backbone coupled with a diffusion prior~\cite{ramesh2022hierarchical}, alongside the MixCo~\cite{kim2020mixco} contrastive learning loss during training.
The MLP backbone includes a ridge regression module and a Residual MLP module. The ridge regression module maps fMRI data to a lower dimension, and the Residual MLP module further refines the representation in an enhanced hidden space.
Following NeuroClips, we initialize the MLP backbone using a pretrained checkpoint from MindEye2~\cite{scotti2024mindeye2}. 

Since we focus on learning semantics for fMRI embeddings, we 
utilize the MLP backbone as the fMRI encoder.
In the Bottom-Up Semantic Enrichment stage, our training protocol adheres to the default NeuroClips semantic learning settings, incorporating our designed classification and action alignment tasks. In the Top-Down Memory Integration stage, we train the MLP backbone and our newly introduced components: the routing network and the fusion mechanism, and implement LoRA tuning within the Video DiT model.

The output of the MLP backbone serves as fMRI embeddings for alignment, classification, and memory pool retrieval during training.
To match the embedding dimension and token length between the MLP backbone output and the retrieved text embeddings, we 
employ an additional cross-attention layer to integrate the fMRI embeddings into the retrieved text embeddings.
During training, except for hyperparameters specifically noted in NeuroClips, all other settings remain consistent with those outlined in Sec.~4.1 of the main text.
During inference, our method remains straightforward and efficient among different architectures.
It requires only the input of fMRI data to generate decoded video, eliminating the need for complex steps like key frame reconstruction, ControlNet integration, or additional condition generation.

We present quantitative results of our \framework using the MindEye and NeuroClips architectures on the cc2017 dataset, as shown in Tab.~\ref{tab:app_extension}.
Building upon the MindEye and NeuroClips architectures, our method achieves improved semantic-level metrics and comparable spatiotemporal-level metrics, indicating the scalability of our approach across different architectures.

\subsection{Expanding the Memory Pool}\label{app:expand_memory_pool}
To further validate the effectiveness and robustness of our proposed Mixture-of-Memories, we ablate the memory pool size by both reducing its capacity and expanding it using external data. The expansion strategy aims to simulate a more comprehensive cognitive process akin to the human brain, which utilizes both established memories and broader external knowledge to better understand and learn from current stimuli.

To achieve this, we incorporate additional datasets including Animal Kingdom~\cite{ng2022animal}, Dream-1K~\cite{wang2024tarsier}, and BOLDMoments~\cite{lahner2023bold}. We process their videos into short clips, randomly selecting 665 animal videos from Animal Kingdom and 200 human videos from Dream-1K to create an expanded pool of 5185 videos. We then further scale the pool to 6185 videos using the BOLDMoments dataset. For all external data, we extract corresponding text, image, and action embeddings for retrieval.

The quantitative results in Tab.~\ref{tab:app_ablation_expand_pool} demonstrate the stability and scalability of our method across the semantic, pixel, and spatiotemporal levels. Notably, a 50\% reduction in memory size (to 2160 videos) results in only minor degradation (e.g., a 0.1\% drop in Acc$_2$), highlighting the robustness of the retrieval mechanism. Conversely, larger pools consistently boost performance across the metrics.

\begin{figure*}[t]
  \centering
  \includegraphics[width=1.0\linewidth]{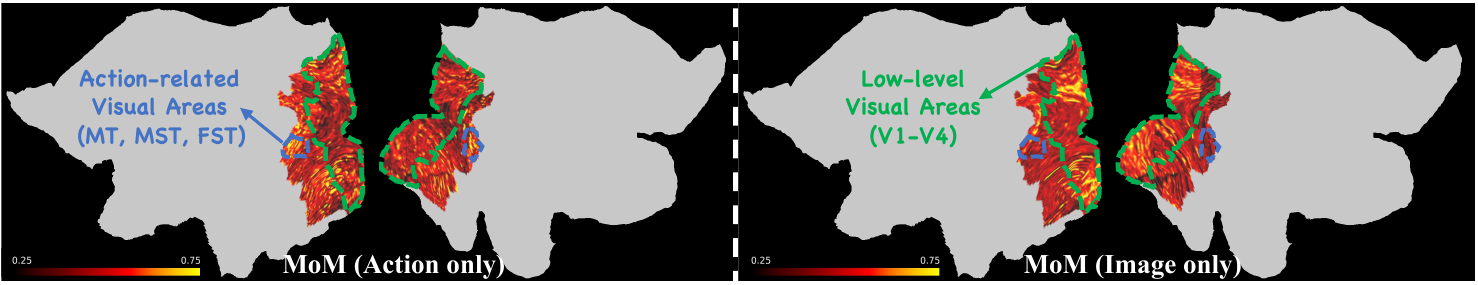}
  \caption{\textbf{More Visualization of voxel weights from the first regression layer.} 
  Voxel weights are averaged and normalized to $[0, 1]$, displayed with a 0.25 to 0.75 colorbar.
  The blue and green dotted lines indicate the action-related and low-level visual areas, respectively.}
  \label{fig:more_inter}
\end{figure*}

\begin{table*}[!ht]
\caption{
\textbf{Noise sensitivity analysis.}
}
\small
\centering
\begin{tabular}{ccccccc} 
\toprule
\textbf{Configuration} & \textbf{Acc$_2$} & \textbf{Acc$_{50}$} & \textbf{SSIM} & \textbf{PSNR} & \textbf{DTC} & \textbf{CLIP-pcc} \\ 
\midrule
\midrule
No Noise & 0.846 & 0.237 & 0.376 & 9.474 & 0.959 & 0.973 \\ 
add 10\% Noise & 0.845 & 0.229 & 0.371 & 9.303 & 0.928 & 0.965 \\
add 25\% Noise & 0.843 & 0.225 & 0.365 & 9.228 & 0.923 & 0.963 \\
\bottomrule
\end{tabular}
\label{tab:re_ablation_noise}
\end{table*}

\begin{table*}[!t]
\caption{
\textbf{The averaged top-1 retrieval accuracy on the cc2017 dataset.}
}
\small
\centering
    \begin{tabular}{c|c|c}
    \toprule
     \textsc{Method} & fMRI-to-image Retrieval ($\uparrow$) &
    image-to-fMRI Retrieval ($\uparrow$) \\
    \midrule
    \midrule
    NeuroClips~\cite{gong2024neuroclips} & 22.2\% & 18.8\% \\ 
    \framework & \textbf{28.3\%} & \textbf{26.2\%} \\ 
    \bottomrule
    \end{tabular}
\label{tab:app_retrieval_acc}
\end{table*}

\subsection{More Visualization Results}\label{app:more_visualization}
\begin{figure*}[p]
  \centering
  \includegraphics[width=0.9\linewidth]{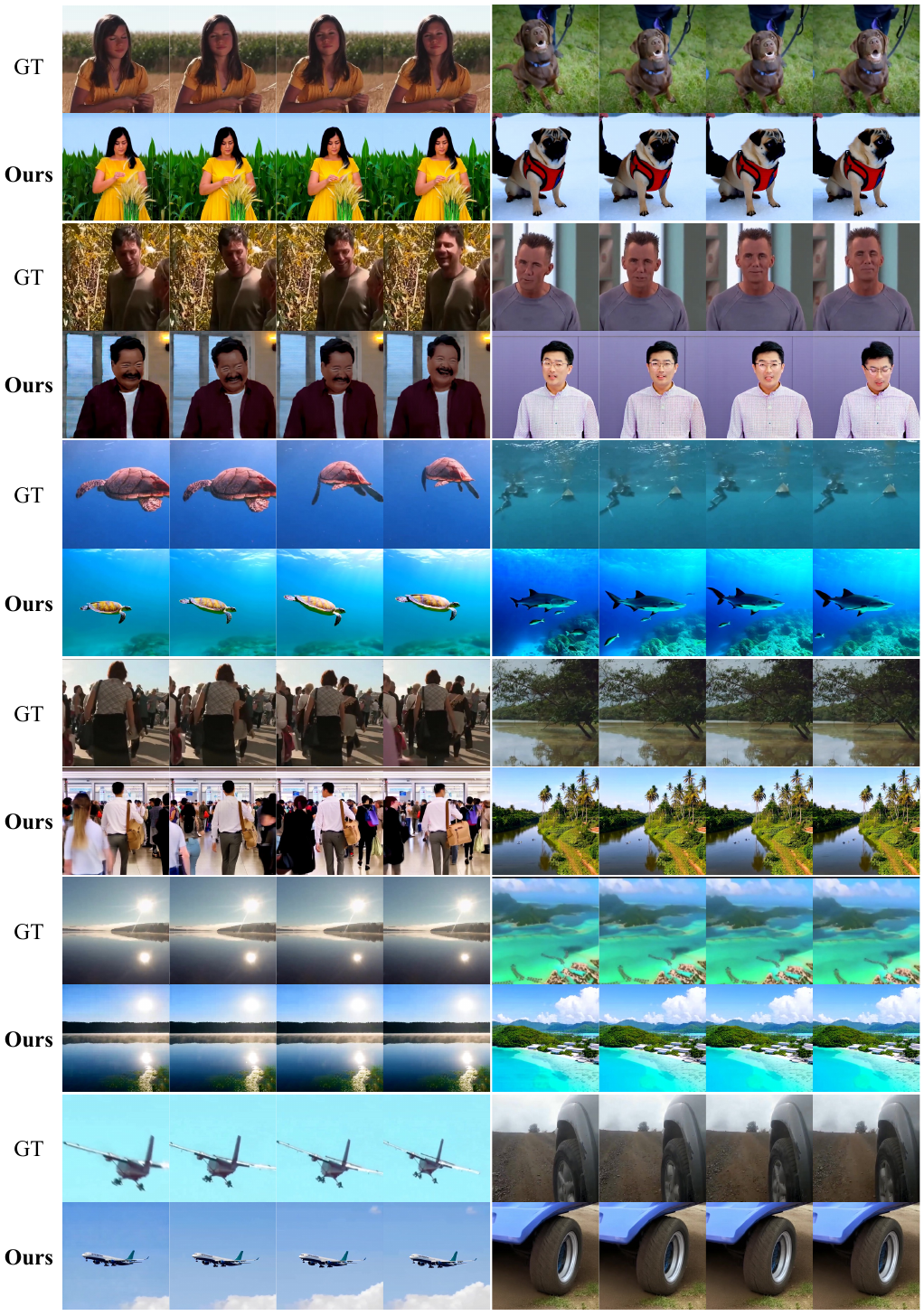}
  \caption{\textbf{More qualitative results of our \framework on the cc2017 dataset.}}
  \label{fig:more_results_wen}
\end{figure*}

\begin{figure*}[h]
  \centering
  \includegraphics[width=1.0\linewidth]{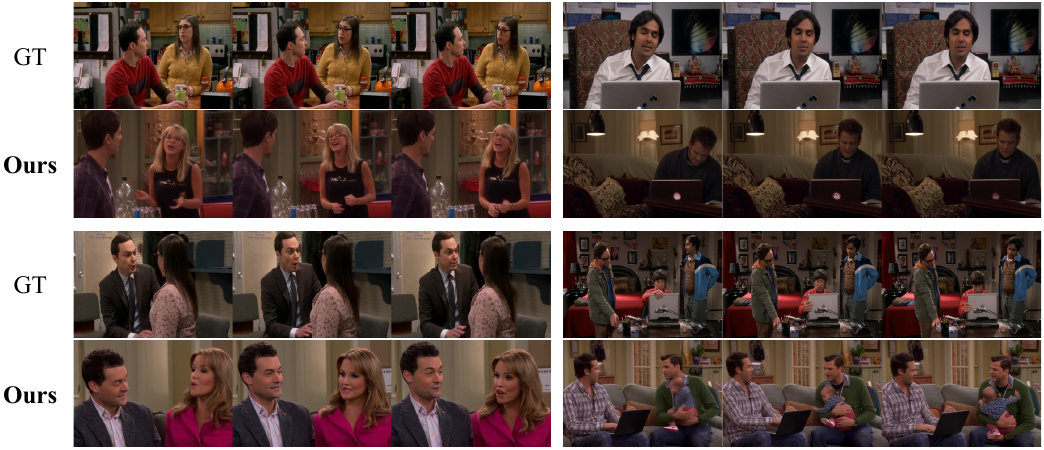}
  \caption{\textbf{More qualitative results of our \framework on the CineBrain dataset.}}
  \label{fig:more_results_cinebrain}
\end{figure*}

To demonstrate the effectiveness of our \framework, we present additional qualitative results on the cc2017 and CineBrain datasets, displayed in Figs.~\ref{fig:more_results_wen} and~\ref{fig:more_results_cinebrain}.

\subsection{Generalization to OOD Concepts}
\label{app:ood}

Our MoM's multi-modal design ensures robustness for OOD samples: if specific image concepts are missing, retrieved text or action priors still enhance generation.
For OOD evaluation, we identify 63 unseen videos based on cc2017 categories and evaluate methods on this subset. Tab.~\ref{tab:re_comparison_ood} shows our method leads on most metrics, indicating strong OOD generalization.
Tab.~\ref{tab:app_ablation_expand_pool} also shows consistent gains with larger pools. This validates our scalability and offers a unique solution to OOD issues by simply increasing memory diversity, a capability absent in other methods.

\subsection{Results on BOLDMoments Dataset}
\label{app:boldmoments}

We report new results on the BOLDMoments~\cite{lahner2023bold} dataset (Version B, MNI152). 
Tab.~\ref{tab:re_comparison_bold} shows that our method consistently outperforms NeuroClips on BOLDMoments, indicating strong cross-dataset generalization.
Tab.~\ref{tab:app_ablation_expand_pool} also shows that adding BOLDMoments data into the memory pool further boosts performance, demonstrating our scalability with increased memory diversity.

\subsection{More Interpretation Results}\label{app:more_interpretation}

We conduct more interpretability analysis in Fig.~\ref{fig:more_inter}, which reveals that: (\textbf{i}) incorporating action memory selectively activates dorsal motion regions (\eg, MT); and (\textbf{ii}) incorporating image memory emphasizes low-level visual areas (\eg, V1-V4). These findings show that MoM aligns with genuine neural processing and our method effectively enhances both motion perception and low-level visual details.

\subsection{Noise Sensitivity Analysis}\label{app:ablation_noise} 
Tab.~\ref{tab:re_ablation_noise} shows that our method remains robust under up to 25\% noise, verifying that MoM provides robust semantic guidance.

\subsection{Retrieval Accuracy}\label{app:retrieval_acc}
Following NeuroClips~\cite{gong2024neuroclips}, we evaluate the retrieval performance of our \framework using Top-1 fMRI-to-image retrieval accuracy (forward retrieval accuracy) and Top-1 image-to-fMRI retrieval accuracy (backward retrieval accuracy).
For fMRI-to-image retrieval, each test fMRI data is converted into an fMRI embedding.
The fMRI embedding is used to query the target embedding based on the CLIP cosine similarity from a set that includes the target embedding along with 299 other randomly selected test embeddings.
Retrieval is successful if the cosine similarity is highest between the fMRI embedding and its corresponding image embedding.
The test set comprises 1,200 fMRI-video pairs, divided into 4 subsets of 300 pairs each for evaluation.
We report the average retrieval accuracy across these subsets.
Image-to-fMRI retrieval follows the same protocol, with fMRI and image roles reversed.

Tab.~\ref{tab:app_retrieval_acc} presents the retrieval accuracy comparison with NeuroClips, demonstrating improvements of 6.1\% in fMRI-to-image retrieval and 7.4\% in image-to-fMRI retrieval.
These results indicate that our proposed method effectively enhances retrieval accuracy, further validating the efficacy of our Mixture-of-Memories strategy.

Despite recent advancements in fMRI-to-image reconstruction achieving retrieval accuracies exceeding 90\%, both NeuroClips and our retrieval accuracy in the cc2017 dataset are lower.
This discrepancy is attributed to three factors:
(1) Task difficulty: Visual stimuli from videos are inherently more complex than images, making the retrieval more challenging.
(2) Dataset distribution: The cc2017 test set includes numerous object categories absent in the training set, increasing generalization difficulty.
(3) Lack of large-scale pretrained models: Unlike the fMRI-to-image reconstruction task, which benefits from robust models pre-trained on large-scale datasets, the fMRI-to-video reconstruction suffers from smaller datasets and lacks pretrained models, complicating the retrieval of unseen samples.

Notably, our approach does not solely depend on retrieval accuracy. Instead, it employs a dynamic and end-to-end retrieval and fusion process, allowing the model to continuously refine the fMRI embeddings and their associated semantics during training. Because the retrieved samples are not fixed but updated based on the current representation, the model can iteratively adjust and improve its alignment between fMRI signals and semantic space.

Future research should focus on constructing large-scale fMRI-video datasets and developing pretrained models to address these issues and further enhance retrieval accuracy.

\subsection{Classification Accuracy}\label{app:cls_acc}
We present our classification accuracy results for each category on the cc2017 dataset, as shown in Tab.~\ref{tab:app_class_acc}.
The results demonstrate that utilizing superclasses and reducing the number of classes, along with the introduction of Focal Loss, effectively enhance classification performance.

\begin{table*}[!ht]
\caption{\textbf{Quantitative ablation of adding EEG input on Subject 5 of CineBrain dataset.}}
\label{tab:ablation_eeg_input}
\centering
\small
    \begin{tabular}{l|cc|ccc}
    \toprule
    \multicolumn{1}{c|}{\multirow{2}{*}{\textsc{Methods}}} & \multicolumn{2}{c|}{Semantic-level} & \multicolumn{3}{c}{Spatiotemporal-level} \\
    \cmidrule(lr){2-3} \cmidrule(lr){4-6}
    & 2-way ($\uparrow$) & 50-way ($\uparrow$) & CLIP-pcc ($\uparrow$) & DTC ($\uparrow$) & MS ($\uparrow$) \\
    \midrule
    \midrule
    only fMRI data & 0.938\scriptsize{$\pm$0.01} & 0.401\scriptsize{$\pm$0.02} & 0.990\scriptsize{$\pm$0.01} & 0.978\scriptsize{$\pm$0.01} & 0.976\scriptsize{$\pm$0.01} \\
    \rowcolor{cyan!10}
    fMRI data + EEG data & \textbf{0.949\scriptsize{$\pm$0.01}} & \textbf{0.471\scriptsize{$\pm$0.04}} & \textbf{0.995\scriptsize{$\pm$0.01}} & \textbf{0.984\scriptsize{$\pm$0.01}} & \textbf{0.979\scriptsize{$\pm$0.01}} \\
    \bottomrule
    \end{tabular}
\end{table*}

\begin{table*}[!ht]
\caption{
\textbf{Ablation on use of superclasses.}
}
\small
\centering
\begin{tabular}{ccccccc} 
\toprule
\textbf{Configuration} & \textbf{Acc$_2$} & \textbf{Acc$_{50}$} & \textbf{SSIM} & \textbf{PSNR} & \textbf{DTC} & \textbf{CLIP-pcc} \\ 
\midrule
\midrule
w/ Subclasses & 0.841 & 0.217 & 0.342 & 8.945 & 0.957 & 0.972 \\ 
\rowcolor{cyan!10} w/ Superclasses (\textbf{Ours}) & \textbf{0.846} & \textbf{0.237} & \textbf{0.376} & \textbf{9.474} & \textbf{0.959} & \textbf{0.973} \\ 
\bottomrule
\end{tabular}
\label{tab:re_ablation_superclasses}
\end{table*}

\begin{table*}[!t]
\centering
\small
\caption{\textbf{Ablation on category and action semantics.}}
\label{tab:re_ablation_semantics}
\begin{tabular}{l|cccc}
\toprule
\textbf{Method} & \textbf{Temp. Cons.} $\uparrow$ & \textbf{EPE} $\downarrow$ & \textbf{Acc$_{2}$} $\uparrow$ & \textbf{Acc$_{50}$} $\uparrow$ \\ \midrule
Category Only & 0.965 & 1.861 & \underline{0.840}  & \underline{0.229} \\
Action Only   & \underline{0.972} & \textbf{1.620} & 0.836 & 0.212 \\
\rowcolor{cyan!10} \textbf{Full Model} & \textbf{0.973} & \underline{1.628} & \textbf{0.846} & \textbf{0.237} \\
\bottomrule
\end{tabular}%
\end{table*}

\begin{table*}[!t]
\caption{
\textbf{Ablation on MoM.}
}
\small
\centering
\begin{tabular}{ccccccc} 
\toprule
\textbf{Configuration} & \textbf{Acc$_2$} & \textbf{Acc$_{50}$} & \textbf{SSIM} & \textbf{PSNR} & \textbf{DTC} & \textbf{CLIP-pcc} \\ 
\midrule
\midrule
MoM w/o Image & 0.842 & 0.225 & 0.344 & 9.274 & 0.950 & 0.970 \\
MoM w/o Action & 0.840 & 0.229 & 0.369 & 9.339 & 0.933 & 0.965 \\ 
\rowcolor{cyan!10} \textbf{Ours} & \textbf{0.846} & \textbf{0.237} & \textbf{0.376} & \textbf{9.474} & \textbf{0.959} & \textbf{0.973} \\ 
\bottomrule
\end{tabular}
\label{tab:re_ablation_mom}
\end{table*}

\section{More Ablation Studies}\label{app:more_ablation}

In this section, we provide more ablation studies and further analysis, investigating the impact of EEG input (Sec.~\ref{app:ablation_eeg}), superclass pre-processing (Sec.~\ref{app:ablation_superclasses}), and different semantic alignments (Sec.~\ref{app:ablation_semantics}).

\subsection{Ablation of EEG Input}\label{app:ablation_eeg}
While fMRI can probe deep-brain neural activities, it is limited by relatively low temporal resolution. In contrast, EEG provides superior temporal resolution that is well-suited for capturing rapid neural oscillations. Combining these two modalities is a promising way to leverage the high temporal resolution of EEG to compensate for the temporal limitations of fMRI.

We conduct preliminary experiments on the CineBrain dataset using both fMRI and EEG data as inputs. As shown in Tab.~\ref{tab:ablation_eeg_input}, incorporating EEG data further improves both semantic and spatiotemporal decoding performance compared with using fMRI alone. These results demonstrate the benefit of integrating EEG with fMRI and suggest that jointly modeling multimodal neural signals is a promising direction for future work.

\subsection{Ablation on Superclass Pre-processing}\label{app:ablation_superclasses}
Tab.~\ref{tab:re_ablation_superclasses} shows that finer-grained categories (w/ Subclasses) hurt both semantic and pixel-level metrics, confirming the necessity of superclass preprocessing for noisy fMRI to reduce learning difficulty.

\subsection{Ablation on Different Semantics}\label{app:ablation_semantics} 
Tab.~\ref{tab:re_ablation_semantics} shows that, in the first stage (bottom-up semantic enrichment), action alignment yields better Temporal Consistency and EPE than category alignment, highlighting its importance for motion decoding, whereas category alignment primarily benefits semantic accuracy.

Furthermore, Tab.~\ref{tab:re_ablation_mom} shows that, in the second stage (top-down memory integration), image and action memories play complementary roles in MoM: images improve low-level details (SSIM) while actions enhance spatiotemporal consistency (DTC).

\section{Limitations}
\label{sec:limitation}
While \frameworkplain has achieved semantically-enhanced and high-quality fMRI-to-video reconstruction, certain limitations remain. Our method struggles with accurately reconstructing cross-scene fMRI, \ie, fMRI recorded during transitions between video clips, a challenge also noted in NeuroClips~\cite{gong2024neuroclips}. Although such fMRI instances are infrequent, addressing this issue is a potential avenue for future research. 
Additionally, due to the limited dataset size, we employed parameter-efficient LoRA for fine-tuning. With sufficient data, fully fine-tuning the model might enhance performance further. Constructing large-scale fMRI-to-video datasets will require collaboration across various fields, including neuroscience and artificial intelligence, to promote the
future research in the community.

\section{Ethical Considerations and Social Impacts}\label{app:impacts}
This work investigates the potential of video generation models for decoding human brain activity, specifically focusing on fMRI data.
This approach aims to enhance our understanding of brain function and contribute to advancements in neuroscience, such as the field of brain-computer interfaces.
While the research holds practical significance, addressing concerns regarding participant privacy and data security is also essential. 
In this work, we utilize two public, de-identified datasets as our training data, thereby strictly adhering to ethical standards.
To further reduce privacy risks, data collection agencies must adhere to stringent protocols and ethical guidelines. Additionally, the community and government should implement measures to safeguard private data and prevent misuse.

\clearpage

\begin{table*}[h]
\caption{
\textbf{Classification accuracy of all categories on the cc2017 dataset.} ``-'' denotes no such category in the test set.
}
\small
\centering
\begin{tabular}{c|c|c}
\toprule
Index & Class Name & Accuracy \\
\midrule
0 & accessory & 0.975 \\
1 & animal & 0.720 \\
2 & appliance & - \\
3 & electronic & 0.967 \\
4 & food & 0.983 \\
5 & furniture & 0.873 \\
6 & indoor & 0.749 \\
7 & kitchen & 0.972 \\
8 & man & 0.718 \\
9 & others & 0.972 \\
10 & outdoor & 0.616 \\
11 & crowd & 0.673 \\
12 & sports & 0.967 \\
13 & vehicle & 0.787 \\
14 & woman & 0.820 \\
\bottomrule
\end{tabular}
\label{tab:app_class_acc}
\end{table*}

\end{document}